\documentclass[runningheads]{llncs}

\usepackage{eccv}

\usepackage{eccvabbrv}

\usepackage{graphicx}
\usepackage{booktabs}

\usepackage[accsupp]{axessibility}  %

\usepackage{hyperref}

\usepackage{orcidlink}

\usepackage{url}
\usepackage{xspace}
\usepackage{booktabs}
\usepackage{epsfig}
\usepackage{graphicx}
\usepackage{tabularx}
\usepackage{colortbl}
\usepackage{multicol}
\usepackage{multirow}
\usepackage{soul}

\usepackage[normalem]{ulem}

\newcommand{\ignorethis}[1]{}
\newcommand{\dataset}{\emph{HierarCaps}\xspace}

\newcommand{\fwk}{RE\xspace}
\newcommand{\framework}{Radial Embedding\xspace}
\newcommand{\frameworks}{Radial Embeddings\xspace}

\newcolumntype{C}{>{\centering\arraybackslash}X}
\newcolumntype{s}{>{\hsize=.02\hsize}X}

\newcommand{\ntrain}{73K\xspace}
\newcommand{\ntest}{1K\xspace}

\newcommand{\literal}[1]{\emph{#1}}
\newcommand{\litquot}[1]{\emph{``#1''}}

\newcommand{\xipos}{\Xi_{\mathbf{r}}(\mathbf{e}, \mathbf{e}')}
\newcommand{\xineg}{\Xi_{\mathbf{r}}(\mathbf{e}, \mathbf{e}'')}
\newcommand{\xir}{\Xi_{\mathbf{r}}}
\newcommand{\thre}{\theta_\mathbf{r}(\mathbf{e})}

\newcommand{\dr}{d_{\mathbf{r}}}
\newcommand{\dre}{d_{\mathbf{r}}(\mathbf{e})}

\newcommand{\myinst}{{\lbrack}INST{\rbrack}}
\newcommand{\mysinst}{{\lbrack}/INST{\rbrack}}

\newcommand{\ourloss}{\mathcal{L}_{\fwk{}}}

\newcommand{\thetad}{\theta(d_{\mathbf{r}}(\mathbf{e}))}

\newcommand{\qual}[1]{\scriptsize{\emph{#1}}}

\definecolor{ec_purple}{rgb}{0.69,0.19,0.69}
\definecolor{cre_green}{rgb}{0.15,0.66,0.25}
\definecolor{cre_red}{rgb}{0.84,0.14,0.16}

\newbox\jsavebox
\newcommand{\jsubfig}[2]{%
	\sbox\jsavebox{#1}%
	\parbox[t]{\wd\jsavebox}{\centering\usebox\jsavebox\\#2}%
	}

\begin{document}

\title{Emergent Visual-Semantic Hierarchies in Image-Text Representations}

\author{Morris Alper \and
Hadar Averbuch-Elor
}

\authorrunning{Alper and Averbuch-Elor}

\institute{
Tel Aviv University
}

\maketitle

\begin{abstract}

While recent vision-and-language models (VLMs) like CLIP are a powerful tool for analyzing text and images in a shared semantic space, they do not explicitly model the hierarchical nature of the set of texts which may describe an image. Conversely, existing multimodal hierarchical representation learning methods require costly training from scratch, failing to leverage the knowledge encoded by state-of-the-art multimodal foundation models. In this work, we study the knowledge of existing foundation models, finding that they exhibit emergent understanding of visual-semantic hierarchies despite not being directly trained for this purpose. We propose the \emph{\framework{}} (\fwk{}) framework for probing and optimizing hierarchical understanding, and contribute the \dataset{} dataset, a benchmark facilitating the study of hierarchical knowledge in image--text representations, constructed automatically via large language models. Our results show that foundation VLMs exhibit zero-shot hierarchical understanding, surpassing the performance of prior models explicitly designed for this purpose. Furthermore, we show that foundation models may be better aligned to hierarchical reasoning via a text-only fine-tuning phase, while retaining pretraining knowledge. Our code, data, and trained models are available at the project page: \url{https://hierarcaps.github.io/}.

\end{abstract}
\section{Introduction}
\label{sec:intro}

\begin{figure}
    \centering
    \includegraphics[trim={0cm 3.5cm 0cm 0},clip,width=0.9\linewidth]{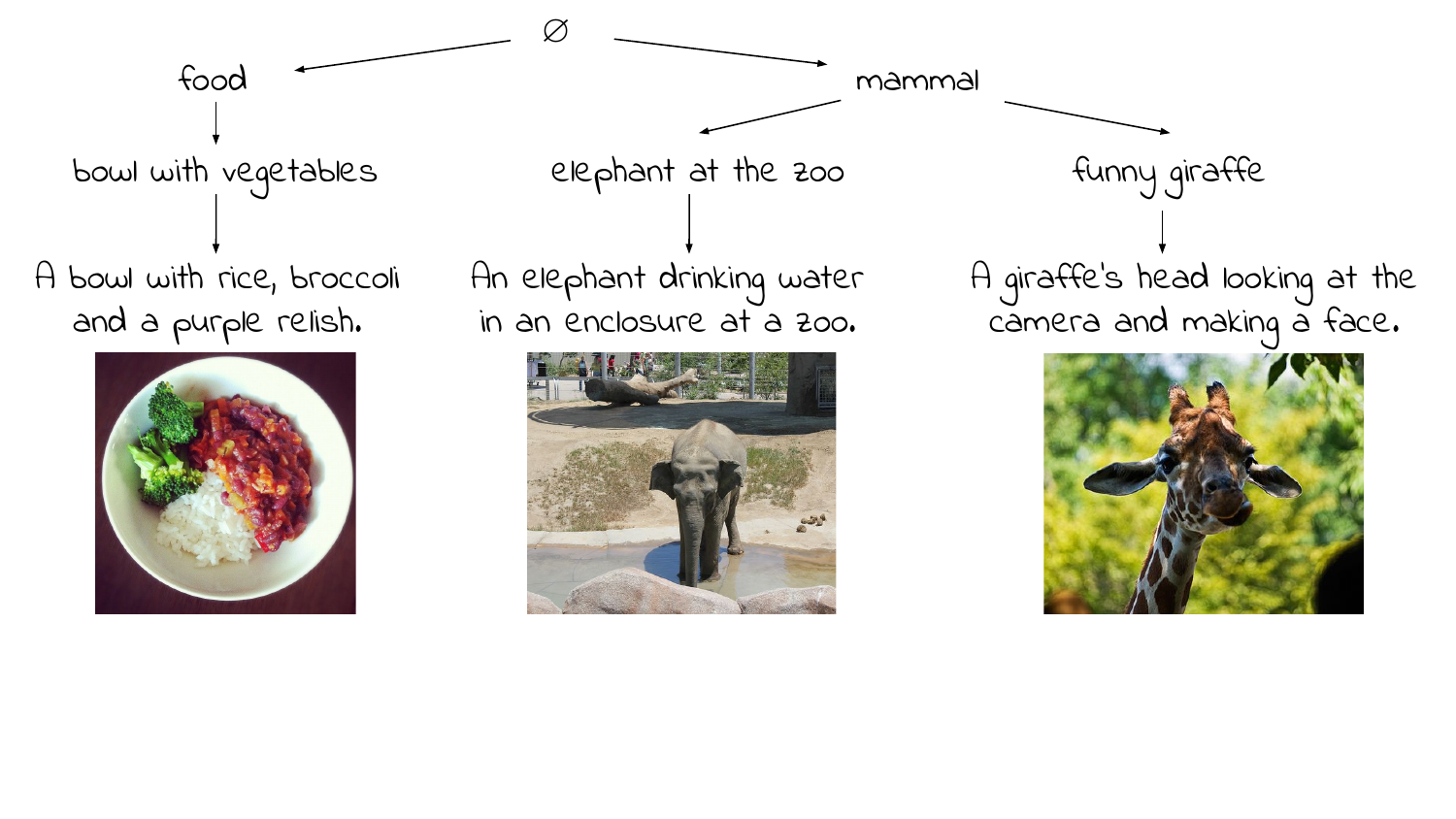}
    \caption{
   A single image may be described by many text of varying levels of descriptiveness. While SOTA multimodal foundation models are commonly used to retrieve a single text matching an image,
   we show that they have learned to model hierarchies.
   By applying our \fwk{} framework to foundation models, we may perform hierarchical image-text matching to
   place images and captions in the context of a \emph{visual-semantic hierarchy} which encompasses the relative meanings of all possible images and texts. 
   Above, we show a slice of the visual-semantic hierarchy obtained with our method, with $\emptyset$ indicating the root node in the hierarchy and arrows corresponding to the logical entailment relation between general and more specific descriptions.
   }
    \label{fig:teaser}
\end{figure}

The visual world is full of hierarchies. Upon seeing something green flying through the sky, the average person will correctly identify it as a parrot, knowing that it is a special bird which is a type of animal. A trained ornithologist might even recognize that it is a ring-necked parakeet. In general, images and their possible descriptions form a \emph{visual-semantic hierarchy}~\cite{vendrov2015order}, a directed graph of logical entailment (implication) between items. As shown in Figure \ref{fig:teaser}, a description such as \literal{mammal} may describe many images, while \literal{funny giraffe} describes a smaller subset. Each image is closest to a highly specific description (e.g. \literal{A giraffe's head looking at the camera and making a face.}) which logically entails the corresponding more general descriptions.

The ubiquitous nature of hierarchies in images suggests that an ideal vision-language model (VLM) would be able to understand them similarly to the aforementioned humans. However, state-of-the-art VLMs such as CLIP~\cite{radford2021learning} were trained on tasks like text-image matching which do not explicitly model hierarchical knowledge. Inspired by prior works studying the knowledge acquired by multimodal foundation models~\cite{alper2023:is-bert-blind,alper2023kiki-or-bouba, zhang-etal-2022-visual,chefer2023hidden,lee2023languageinformed,Trager_2023_ICCV}, we ask the question: Do such VLMs develop an \emph{emergent} understanding of visual-semantic hierarchies? Such understanding could shed light on the inner working of powerful foundation models typically used as black boxes; in addition, prior work has found hierarchical understanding to provide an important inductive bias to improve performance on tasks such as image classification~\cite{bertinetto2020making, li2019large, novack2023chils, phoo2021coarsely, xu2023challenges, yi2022exploring}. We investigate this both in zero-shot as well as fine-tuned settings, to both investigate the presence of this emergent knowledge used as-is as well as its utility as an initialization for transfer learning applied to hierarchical tasks.

In order to probe and optimize models for hierarchical understanding, we must first define the geometry which represents hierarchical entailment in a continuous space. Prior works have represented logical entailment between texts or between images and text as a partial order relation~\cite{vendrov2015order}; in particular, the Entailment Cone (EC) framework represents entailment as inclusion in a cone radiating from a point away from the origin~\cite{ganea2018hyperbolic}. However, these works train models from scratch with EC-based objectives, as well as generally operating in hyperbolic space (while common foundation VLMs operate in Euclidean space). To better model the emergent hierarchical geometry of foundation VLMs, we introduce the \framework{} (\fwk{}) framework, which relaxes the assumptions of EC to match the existing geometric configuration of pretrained VLM representations. We show that the \fwk{} framework demonstrates the emergent hierarchical knowledge of VLMs. Furthermore, we propose a RE-based contrastive objective for model fine-tuning to enhance hierarchical understanding in VLMs, and show that this outperforms EC-based optimization for these models.

A significant obstacle for studying visual-semantic hierarchies is the lack of ground-truth data, as existing datasets only contain images paired with single or unrelated reference captions. Therefore, we propose the \dataset{} dataset, containing 73K images paired with multiple valid texts arranged in a logical hierarchy. To create \dataset{}, we leverage large-scale image-caption Internet data and enrich it with hierarchical text generation using a large language model (LLM) and natural language inference (NLI). We also provide a manually-curated 1K-item test set and contribute evaluation metrics for benchmarking visual-semantic hierarchical understanding, quantifying the multimodal hierarchical knowledge which prior works only evaluate qualitatively~\cite{vendrov2015order,desai2023hyperbolic}.
We show that \dataset{} may supervise VLM fine-tuning with our \fwk{} framework, generally boosting hierarchical understanding of pretrained models. Importantly, this can be achieved while still retaining pretrained knowledge -- thus performing model \emph{alignment}, parallel to the alignment stages of recent foundation models which unlocks emergent abilities such as instruction following~\cite{ouyang2022training,dai2023instructblip}.

We validate the effectiveness of our approach on \dataset{} and on existing benchmarks for related tasks such as lexical entailment prediction and hierarchical classification. Our evaluation shows that foundation VLMs indeed exhibit emergent hierarchical understanding, significantly outperforming prior models which learn hierarchical representations directly even in the zero-shot setting, with further overall performance enhancement after fine-tuning.
We release\footnote{\url{https://hierarcaps.github.io/}} our code, data, and trained models to the research community to spur future progress on multimodal hierarchical understanding.
\section{Related Work}
\label{sec:rw}

\noindent \textbf{V\&L representation learning.} 
While earlier work in image representation learning focused on models trained on supervised benchmarks such as ImageNet~\cite{krizhevsky2012imagenet}, recent approaches have achieved state-of-the-art via contrastive training on web-scale collections of paired images and text\cite{radford2021learning,jia2021scaling}.
These multimodal representations exhibit various emergent geometric properties which were not explicitly optimized.
For instance,
vector arithmetic in CLIP's semantic space has been found to correlate with semantic relationships, similar to well-known properties of word embeddings~\cite{couairon2022embedding,tewel2022zerocap,Trager_2023_ICCV}. Liang et al.~\cite{NEURIPS2022_702f4db7} explore the geometry of text and image embeddings in CLIP space, finding that the modalities are contained in narrow cones separated by a consistent cross-modal gap.
Similarly, our work explores the emergent geometry of hierarchies in V\&L representations. In addition, our text-only alignment approach bears some similarity to works focusing on deficiencies in textual representations of existing VLMs and how to adapt these models to imbue additional textual understanding~\cite{zhai2022lit,yuksekgonul2022and,kamath2023text,reimers2020making,carlsson2022cross,chen2023mclip}.

\smallskip
\noindent \textbf{Hierarchical multimodal reasoning.}
Images may be described by multiple valid texts of varying granularity levels, forming a hierarchy which may be learned or exploited in various ways. One approach infers concept hierarchies from paired text and images, learning text
and image embeddings jointly to implicitly infer the generality of concepts and their logical entailment structure~\cite{kiela2015exploiting,vendrov2015order,zhang2016learning}. Other works generate texts or knowledge graphs from images with varying levels of detail and attention to different regions; methods include controllable captioning~\cite{deng2020length,Chen_2021_CVPR}, contrastive captioning~\cite{dai2017contrastive,luo2018discriminability}, dense captioning~\cite{johnson2016densecap}, guided decoding~\cite{kornblith2023guiding}, and scene graph generation~\cite{zhu2022scene}.
Another approach uses existing hierarchical structure in ground-truth annotations to improve predictions for tasks such as image classification~\cite{li2019large,bertinetto2020making,phoo2021coarsely}; these works use concept hierarchies over categorical image labels, while our benchmark consists free text arranged in entailment hierarchies, not confined to a small set of categories.

We are currently witnessing pioneering efforts in understanding the semantic hierarchies learned by neural VLMs such as CLIP, though current knowledge in this field is still sparse. Xu et al.~\cite{xu2023challenges} find that VLMs are more successful at matching fine-grained concepts with images, while underperforming on general concepts (e.g. \literal{leopard} vs. \literal{feline}). Yi et al.~\cite{yi2022exploring} show that CLIP may be outperformed on ImageNet classification by a model explicitly trained using concept hierarchies from WordNet. Novack et al.~\cite{novack2023chils} propose a zero-shot classification pipeline with CLIP by leveraging concept hierarchies during inference. Desai et al.~\cite{desai2023hyperbolic} introduce neural hyperbolic V\&L embeddings with hierarchical image--text matching as one use case, showing qualitative results alone for this task.

\smallskip
\noindent \textbf{Hierarchical embeddings.}
Hierarchical representations have attracted interest in various fields due to the hierarchical nature of of many types of data. A number of works have approached text (primarily words) as hierarchical for the purpose of learning embeddings to model semantic relationships such as hypernymy~\cite{nguyen2017hierarchical,chang2017distributional,vulic2017specialising,athiwaratkun2018hierarchical,dash2020hypernym}. Many works model data with geometric objects that directly represent relations between items~\cite{xiong2023geometric}, notably including the use of hyperbolic manifolds due to their attractive properties for representing hierarchical graphs~\cite{sarkar2011low, suzuki2019hyperbolic}.
Nickel and Kiela~\cite{nickel2017poincare} introduce hyperbolic word embeddings, and subsequent works refine this idea by exploring different models of hyperbolic space~\cite{nickel2018learning}, the geometry of entailment relationships~\cite{ganea2018hyperbolic}, and extensions to texts of arbitrary length~\cite{dhingra2018embedding,zhang2021hype}. Recent work has also applied hyperbolic learning to various tasks in computer vision~\cite{dhall2020hierarchical,atigh2022hyperbolic,ermolov2022hyperbolic,li2023euclidean,mettes2023hyperbolic} and multimodal learning~\cite{desai2023hyperbolic,long2023cross,hong2023hyperbolic}. While we take inspiration from these works, particularly the hierarchical nature of text-image data from Desai et al.~\cite{desai2023hyperbolic} and the entailment cone framework introduced by Ganea et al.~\cite{ganea2018hyperbolic}, we focus on leveraging the strong knowledge of existing pretrained models such as CLIP which is embedded in Euclidean space, rather than training from scratch in hyperbolic space.
\section{Preliminaries} \label{sec:prelim}
We proceed to cover preliminary mathematical and geometric concepts which provide background for our study of hierarchical knowledge in VLMs. We refer the reader to Ganea et al.~\cite{ganea2018hyperbolic} and Desai et al.~\cite{desai2023hyperbolic} for further exposition and illustration of the concepts described below.

\smallskip
\noindent
\textbf{Notation.} Our discussion includes the following notation: $\operatorname{sim}(\mathbf{v}, \mathbf{w}) := \frac{\mathbf{v} \cdot \mathbf{w}}{\|\mathbf{v}\|\|\mathbf{w}\|}$ denotes the cosine similarity function between two vectors, and the angle spanned by three points $\mathbf{a, b, c}$ in space is denoted by $\angle \mathbf{abc}$. $\left[x\right]^+ := \max(0, x)$ is the positive part (ReLU) function.

\smallskip
\noindent
\textbf{Geometric Preliminaries.}
Consider embeddings in embedding space $M$ (e.g. $\mathbb{R}^n$ for Euclidean embeddings). We are interested in embeddings which represent hierarchical relations between items. In particular, embeddings $\mathbf{e}, \mathbf{e}' \in M$ of items $x, x'$ should have a particular geometric configuration if $x$ is entailed by $x'$. For example, for text embeddings, if $x = $\litquot{animal} and $x' =$\litquot{cat} then $x$ is entailed by $x'$ and thus $\mathbf{e}$ and $\mathbf{e}'$ should have a certain relative positioning in embedding space. Denote by $\mathbf{r} \in M$ the \emph{entailment root} (or simply \emph{root}), a special fixed point in space which is the anchor used as a reference point for all entailment relationships. Considering distinct $\mathbf{e}, \mathbf{e}' \in M$ (also distinct from the root), we define the exterior angle $\xipos := \pi - \angle \mathbf{r} \mathbf{e} \mathbf{e}' = \arccos \left(\operatorname{sim}(\mathbf{e} - \mathbf{r}, \mathbf{e}' - \mathbf{e})\right)$, and $d_{\mathbf{r}}(\mathbf{e}) := \|\mathbf{e} - \mathbf{r}\|$, the Euclidean distance of $\mathbf{e}$ from the root.

\smallskip
\noindent \textbf{Entailment Cone Embeddings.}
Ganea et al.~\cite{ganea2018hyperbolic} introduce the Entailment Cone (EC) framework for hierarchical representation learning. EC embeddings possess a defined geometric relation between items which is a \emph{partial order}, a desirable mathematical property for modeling logical entailment. For instance, partial orders and logical entailment both obey transitivity (e.g. \literal{black cat} is a type of \literal{cat} which is a type of \literal{animal}, so \literal{black cat} is a type of \literal{animal}).  See the supplementary materials for the mathematical definition of partial orders and further explanation of their relevance to entailment hierarchies.

\begin{figure}
    \hfill
    \jsubfig{\includegraphics[trim={0 0 0 0},clip,width=0.4\textwidth]{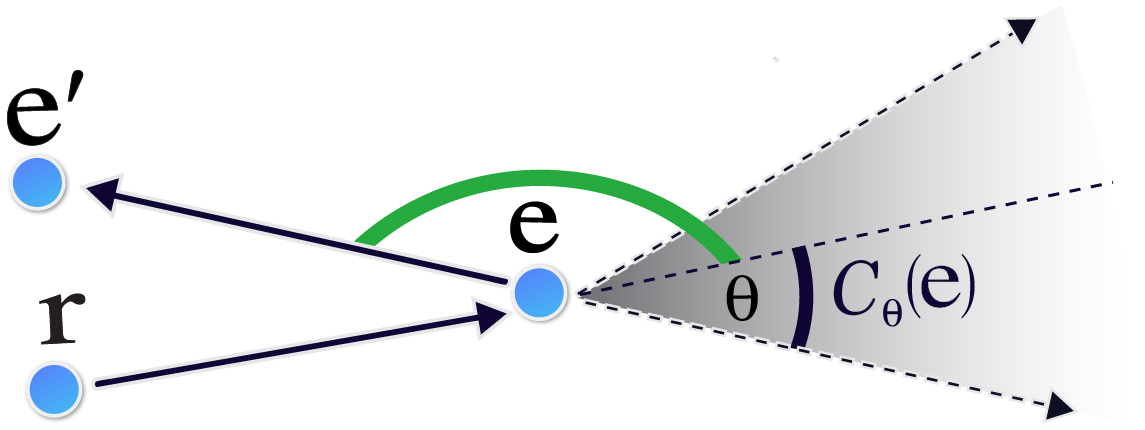}}{
        {\small
            $\mathcal{L}_{EC} = \left[{\color{cre_green} \xipos} - \thre\right]^+$
        }
    }
    \hfill
    \jsubfig{\includegraphics[trim={0 0 0 0},clip,width=0.45\textwidth]{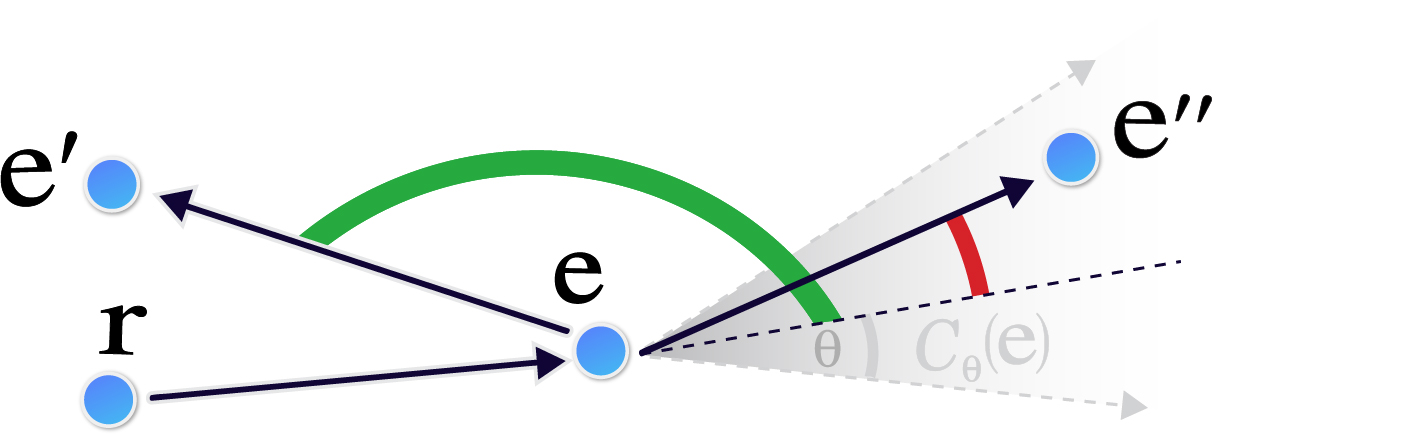}}{
        {\small
            $\ourloss = {\color{cre_green} \xipos}-{\color{cre_red} \xineg}$
        }
    }
    \hfill
    \caption{\textbf{Illustration of EC and \fwk{} optimization.}
    Above we show examples of cases of positive loss under both frameworks.
    In the EC framework (left), any embedding in the cone $C_{\thre}(\mathbf{e})$ represents an item entailed by $\mathbf{e}$ (while $\mathbf{e}'$ is outside this cone). The half-aperture angle $\thre$ varies with distance from the root embedding $\mathbf{r}$ to enforce a partial order. During training, the deviation from this cone defines a margin loss.
    In our proposed \fwk{} framework (right), the loss is instead given by the difference between exterior angles of positive and negative examples, and with no dependency on $\thre$. In the above case, this loss is positive since the positive item $\mathbf{e}'$ has a larger exterior angle than the negative item $\mathbf{e}''$.
    }
    \label{fig:framework}
\end{figure}

In the EC setting, a relation between two embeddings $\mathbf{e}, \mathbf{e}'$ is defined by $\mathbf{e} \leq \mathbf{e}' \leftrightarrow \mathbf{e}' \in C_{\theta}(\mathbf{e})$, where $C_\theta(\mathbf{e}):= \{\mathbf{e}' \in M : \xipos \leq \theta\}$ is a convex cone with half-aperture angle $\theta$ originating at $\mathbf{e}$ and radiating away from the origin. See Figure \ref{fig:framework} for an illustration. Ganea et al. show that this is a partial order if $\theta$ is defined as a function $\thre$ which varies with the distance of $\mathbf{e}$ from the root $\mathbf{r}$; in Euclidean space, as $\sin \thre = \epsilon / \dre$ for constant $\epsilon > 0$.

In the EC framework the positive excess $\mathcal{L}_{EC} := \left[\xipos - \thre\right]^+$ may be used as a margin loss to encourage pairs of inputs with an entailment relation to satisfy the partial order relation~\cite{ganea2018hyperbolic,desai2023hyperbolic}.
For example, as shown in the left-hand side of Figure \ref{fig:framework},
if $(\mathbf{e}, \mathbf{e}')$ are embeddings of a pair of items in an entailment relationship (e.g. \literal{animal} and \literal{cat}), then this loss pushes $\mathbf{e}'$ into the entailment cone $C_{\thre}(\mathbf{e})$ if it deviates from this cone.

\section{\frameworks{}} \label{sec:ours}
We now introduce our \framework{} (\fwk{}) framework, which relaxes the assumptions of EC and modifies its optimization method for compatibility with the representations learned by pretrained foundation VLMs.
Our underlying observation is that foundation VLM representations typically place generic concepts in central locations relative to related, more specific concepts. Intuitively, this agrees with the contrastive objective of models like CLIP which encourages similar concepts to cluster together; for instance, the embedding of \literal{animal} should be close to related, specific concepts such as \emph{dog}, \emph{zebra}, and \emph{fish};
hence, it is expected to be more centrally located. This suggests an emergent hierarchical structure.
To uncover (and optionally enhance) this hierarchical structure, we relax assumptions tying entailment to a partial order relation based on entailment cones, as pretrained foundation VLMs have already learned a configuration in high-dimensional Euclidean space that does not necessarily abide by the strong requirements of EC. Instead, we identify the inherent hierarchical geometric configuration of pretrained VLMs, first locating its natural entailment root and then defining functions for hierarchical understanding.

\smallskip
\noindent
\textbf{Empty String as Entailment Root} \label{sec:empty}
To identify an entailment root, we exploit the ability of VLMs to encode generic semantics in underspecified inputs. To this aim, we use the embedding of the empty string $\mathbf{e}_\emptyset$ as the entailment root $\mathbf{r}$, as an empty caption may accompany virtually any image. This allows us to use the pretraining knowledge encapsulated by these models to match their learned embedding configurations. Furthermore, when we perform model alignment (described below) this embedding affects the gradient of the model's weights as part of the loss function, thus optimizing its position.

\smallskip \noindent
\textbf{Measuring Genericness and Entailment with \fwk{}.}
Given the entailment root $\mathbf{r}$ defined above, we adopt the use of the function $\dr(\cdot)$ to measure concept genericness~\cite{desai2023hyperbolic}. To measure entailment, we use the exterior angle function $\xir(\cdot, \cdot)$; unlike prior work on entailment cones, we use its value directly without reference to an absolute threshold. This allows us to exploit hierarchical structure in embeddings without making strong assumptions such as the existence of an EC-based partial order.

\subsection{\fwk{}-Based VLM Alignment}
\label{sec:our_loss}

In order to fine-tune pretrained VLMs for enhanced hierarchical understanding, we propose a contrastive objective which separates positive and negative pairs without requiring an absolute point of reference for entailment. We exploit the configuration of hierarchical embeddings and the design of \dataset{} (as described in Section \ref{sec:dataset}) by assuming that for each positive pair ($\mathbf{e}, \mathbf{e}')$ there is a negative pair $(\mathbf{e}, \mathbf{e}'')$. For example, $(\mathbf{e}, \mathbf{e}', \mathbf{e}'')$ could be the embeddings of the texts $t=$\litquot{animal}, $t'=$\litquot{goat}, and $t''=$\litquot{portrait} respectively, as $t$ is entailed by $t'$ and $t$ contradicts $t''$ (and thus is not entailed by $t''$). Hence, we define the contrastive \fwk{} loss function $\ourloss := \xipos - \xineg$.

Conceptually, $\ourloss$ encourages the positive pair $(\mathbf{e}, \mathbf{e}')$ to have a relatively smaller exterior angle than the negative pair $(\mathbf{e}, \mathbf{e}'')$, as illustrated in Figure \ref{fig:framework}. A direct comparison  to $\mathcal{L}_{EC}$ shows significant performance gains for hierarchical image--text retrieval (as illustrated in Section \ref{sec:ablations}). In the supplementary material, we also show that $\ourloss$ can be derived as the limit of a margin EC loss applied to contrastive pairs as the margin tends to infinity, which empirically outperforms smaller margins in our evaluation.

We emphasize that $\ourloss$ is fundamentally different from conventional contrastive loss functions. While standard triplet losses also consider triplets of items such as $(\mathbf{e}, \mathbf{e}', \mathbf{e}')$, in the standard case the positive relation between $\mathbf{e}$ and $\mathbf{e}'$ is symmetric~\cite{chuang2020debiased, shah2022max}. By contrast, in our case $\mathbf{e}$ and $\mathbf{e}'$ cannot be swapped, because logical entailment is an asymmetric relation. This is seen in the above example where \litquot{goat} is a type of animal while \litquot{animal} is not a type of goat. Hence, the exterior angle terms $\xir(\mathbf{e}, \cdot)$ in the definition of $\ourloss$ are asymmetric functions of their arguments, unlike the Euclidean or cosine distance functions typically used in contrastive learning. Moreover, $\ourloss$ depends on the learnable root $\mathbf{r}$. This loss pushes items entailed by $\mathbf{e}$ towards the half-line $(\mathbf{r}, \mathbf{e})$, unlike standard contrastive losses which only encourage similar items to be close and dissimilar items to be far apart in representation space.
\section{The \dataset{} Dataset}
\label{sec:dataset}

While there exist large-scale datasets of paired images and captions, these typically contain a single caption per image or multiple independent reference texts. To evaluate and optimize visual-semantic hierarchical understanding, we propose a new dataset and benchmark, \dataset{}, containing images paired with multiple valid texts arranged in a logical hierarchy. To generate this data we use existing image captioning datasets along with a LLM- and NLI-guided hierarchy generation procedure; the train set is produced fully automatically, while the test set is manually curated and corrected to serve as a clean evaluation benchmark.

\begin{figure}
    \begin{minipage}{0.18\linewidth}
        \includegraphics[width=\linewidth]{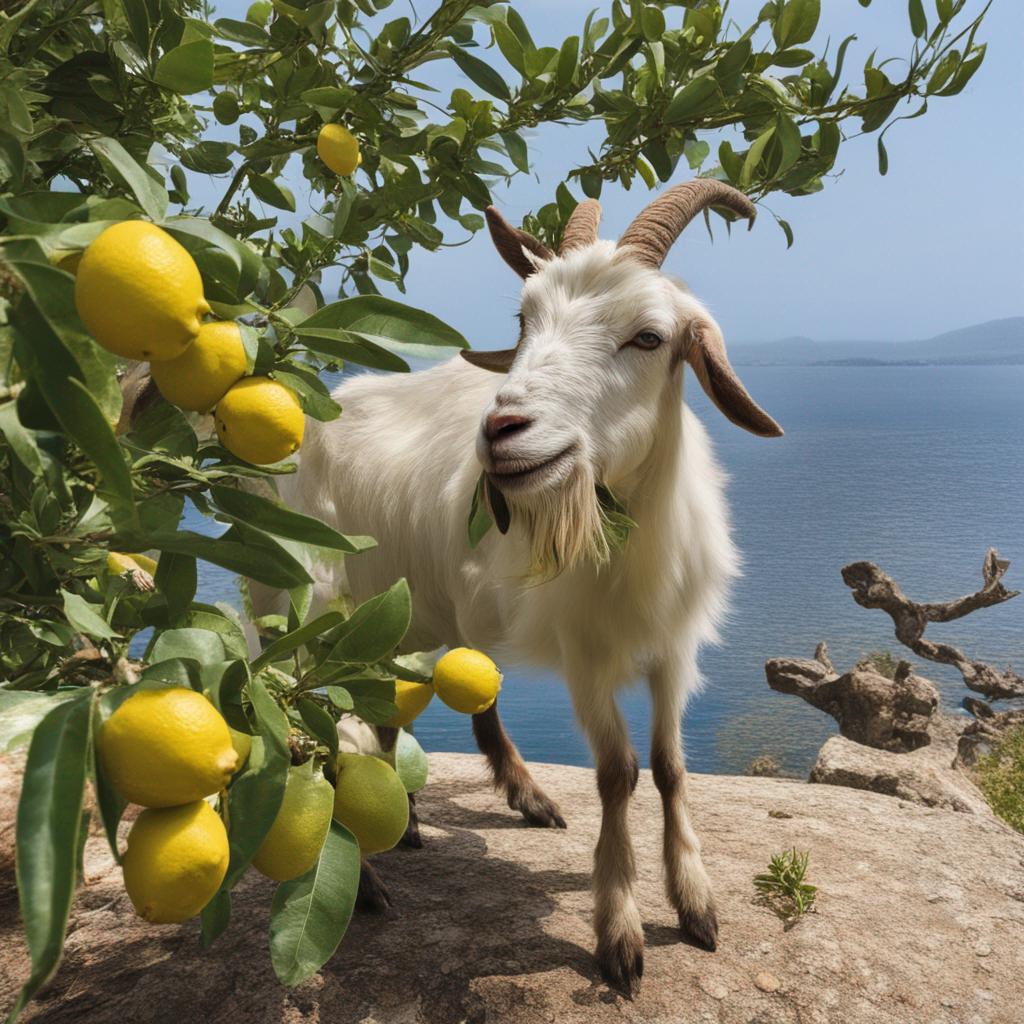}
    \end{minipage}
    \hfill
    \begin{minipage}{0.78\linewidth}
        \begin{tabularx}{\linewidth}[t]{CC}
            {\scriptsize \textbf{Positive}} & {\scriptsize \textbf{Negative}}\\
            \cellcolor{OliveGreen!10} \qual{animal} & \cellcolor{red!10}
            \qual{portrait} \\
            \cellcolor{OliveGreen!20} \qual{goat} & \cellcolor{red!20} \qual{frog} \\
            \cellcolor{OliveGreen!30}
            \qual{goat on island} & \cellcolor{red!30}
            \qual{goats graze at snowy night} \\
            \cellcolor{OliveGreen!40}
            \qual{a goat eating leaves of a lemon tree on the island
            } & \cellcolor{red!40}
            \qual{goat on island with flowers blooming in the spring}
        \end{tabularx}
    \end{minipage}
    \caption{\textbf{Sample item from the \dataset{} train set}\setcounter{footnote}{0}\protect\footnotemark{}. Ground-truth captions have a four-tiered hierarchical structure. The first tier contains the most generic description matching the image (\literal{animal}), the last contains the most specific description (\literal{a goat eating leaves...}), and each (positive) tier is logically entailed by the following tier. The train set also contains corresponding negative captions; corresponding captions in the same tier logically contradict each other, and each positive caption is implied by both (positive and negative) captions in the following tier.}
    \label{fig:dataset_train}
\end{figure}

\subsection{Design and Contents}
\label{sec:dataset_design}

\footnotetext{We show a synthetic image in place of \href{https://c8.alamy.com/comp/F4M5CE/a-goat-eating-leaves-of-a-lemon-tree-on-the-greek-island-of-zakynthos-F4M5CE.jpg}{the original image} from Conceptual Captions due to licensing.} 

\dataset{} contains images with paired hierarchical caption data, including a \ntrain{}-item train set and \ntest{}-item test set. See Figure \ref{fig:dataset_train} for a sample from its train set.
Items in \dataset{} have a four-tiered hierarchical structure, designed by supplementing ground-truth Internet image captions with a minimal amount of added data to encompass the tasks of lexical and textual entailment.
\emph{Lexical entailment} refers to the logical relation between words such as \emph{cat} implying \emph{animal} (as a cat is a type of animal), while \emph{textual entailment} refers to logical implication between full texts, such as \literal{this is a large cat} implying \literal{this is a cat}. As seen in Figure \ref{fig:dataset_train}, the first two tiers in \dataset{} caption hierarchies roughly correspond to lexical entailment, as they usually contain single words or short phrases, while the last two tiers correspond to textual entailment of longer texts. In this way, training or evaluating with \dataset{} subsumes hierarchical understanding on the two varieties of entailment, which have both been of significant interest in NLP and multimodal learning.

Each image has four accompanying positive captions (seen on the left of Figure \ref{fig:dataset_train}). The text in the fourth tier (most specific) is the original caption from an image-caption dataset, while all other texts are LM-generated (described in in Section \ref{sec:dataset_construction}), forming a logical entailment hierarchy (e.g. \literal{animal} $\rightarrow$ \literal{goat} $\rightarrow$ \literal{goat on island} $\rightarrow$ \literal{a goat eating leaves of a lemon tree on the island}). In addition, images in the train set also possess four counterfactual negative captions (seen on the right of Figure \ref{fig:dataset_train}). These are constructed to logically contradict the corresponding positive item in the same tier while also entailing the positive item from the previous tier; for example  \literal{frog} in tier 2 of the figure entails \literal{animal}, while contradicting \literal{goat}. When using \dataset{} for fine-tuning, the positive and negative from each tier provide the paired contrastive data needed for the \fwk{} framework's loss function, as described in Section \ref{sec:our_loss}.

\begin{figure}
    \jsubfig{\includegraphics[trim={0 0 9.8cm 0},clip,height=2.5cm]{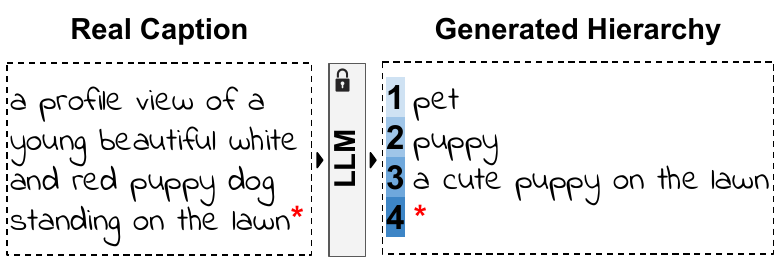}}{{\small Seed Hierarchy Generation}}
    \hfill
    \jsubfig{\includegraphics[trim={15cm 0 0 0},clip,height=2.5cm]{figures/pipeline/pipe.pdf}}{{\small Hierarchy Completion}}
    \caption{\textbf{Dataset construction pipeline}, used to create \dataset{}. Real image captions from Conceptual Captions are fed to a LLM with various prompts to produce caption hierarchies; above, \textcolor{red}{*} indicates the original caption, enriched with a full hierarchy of shorter, logically entailed captions. These are filtered with an NLI model to enforce logical entailment and augmented with few-shot LLM text completion, producing a set of \emph{seed hierarchies}. We then distill these hierarchies into a much smaller language model $\mathcal{H}$ which learns to complete hierarchies in both directions; as described in Section \ref{sec:dataset_construction}, $\mathcal{H}$ is used to produce the final hierarchies in \dataset{}.
    }
    \label{fig:pipeline}
\end{figure}

\subsection{Dataset Construction}
\label{sec:dataset_construction}

Modern multimodal learning benefits from large-scale data, particularly an abundance of paired images and textual captions. Unfortunately, in our setting the desired data (multiple hierarchically-arranged captions for an image) does not naturally occur alongside Internet images. Images on the Internet most commonly appear with a single caption (or alt text); this data was used for example to produce the Conceptual Captions~\cite{sharma2018conceptual} (CC) dataset. More reference texts per image may be produced with manual annotation, as was done for the Microsoft COCO Captions~\cite{lin2014microsoft,chen2015microsoft} dataset, but in this case each reference is produced independently and thus they do not form a consistent hierarchical structure. Therefore, we instead build upon these existing image-caption datasets to enrich them with hierarchical texts, using the powerful text generation abilities of modern LLMs and logical understanding of pretrained NLI models.

Our dataset construction pipeline is illustrated in Figure \ref{fig:pipeline}. We first automatically create seed data using captions from the train set of CC, using the SOTA LLM Llama 2~\cite{touvron2023llama} for text generation along with heavy NLI filtering to enforce logical entailment. We engineer prompts to produce the desired output for each tier; these prompts are reproduced in the supplementary material. While this process is inefficient and only succeeds in a minority of cases, we augment them with more synthetic hierarchies using in-context completion with Llama 2. We then distill this core of examples into a smaller language model $\mathcal{H}$ (using the encoder-decoder model Flan-T5~\cite{chung2022scaling}) that learns to generate a valid positive hierarchy given an input detailed caption. We then run it on all input captions (CC train captions for our train set, COCO validation captions for our test set). $\mathcal{H}$ is trained to complete caption hierarchies in both directions (specific to general and general to specific); on the train set, after producing the positive hierarchies we run $\mathcal{H}$ in the general$\to$specific direction to deliberately hallucinate negative captions. All outputs are filtered with NLI to ensure the correct entailment and contradiction relations between texts.

Along with \ntrain{} automatically-generated train items, we manually review and correct \ntest{} test items as a clean evaluation benchmark. When performing this test set curation, we found 75\% of automatically-generated items to be fully valid; remaining items typically contained minor inaccuracies. As we show below, training on our fully automatically-generated train set provides an empirical performance boost despite the presence of such noise.  See the supplementary material for further dataset construction details, including models used, inference settings, examples of inaccuracies produced by automatic generation, and additional technical details.
\section{Experiments}
\label{sec:exp}

\smallskip
\noindent \textbf{Models Considered.} We focus on dual encoder VLMs (i.e. paired text and image encoders). We evaluate OpenAI and LAION implementations of CLIP~\cite{radford2021learning, cherti2023reproducible} which we refer to as \literal{CLIP} and \literal{OpenCLIP} respectively, and we test an open implementation of ALIGN~\cite{jia2021scaling}. We test various model sizes; see the supplementary material for checkpoints and numbers of parameters.

\smallskip
\noindent \textbf{Full Alignment Framework}.
As seen in Figure \ref{fig:dataset_train}, each caption hierarchy in the \dataset{} train set consists of four positive captions $(P_1, P_2, P_3, P_4)$ and four negative captions $(N_1, N_2, N_3, N_4)$; we use caption triplets $(P_i, P_{i+1}, N_i)$ for tiers $i \in \{1, 2, 3\}$ to calculate the \fwk{} loss $\mathcal{L}_{\fwk}$ as defined in Section \ref{sec:our_loss}. We aggregate this loss over tiers 1--3 along with hard example mining within minibatches to compute aggregate \fwk{} loss $\tilde{\mathcal{L}}_{\fwk}$.
We also add an additional loss term that discourages fine-tuned text embeddings from deviating from their original values. Given fine-tuned embeddings $\{\mathbf{e}_i\}_{i=1,\cdots,8}$ of the image's ground-truth caption texts, and corresponding original (pretrained, not fine-tuned) embeddings $\{\mathbf{e}_i^*\}_{i=1,\cdots,8}$, we define the regularization loss $\mathcal{L}_{reg} := - \frac{1}{8}\sum_{i=1}^8 \operatorname{sim}(\mathbf{e}_i, \mathbf{e}_i^*)$, i.e. the mean cosine similarity loss between pretrained and fine-tuned text representations.
The total loss for fine-tuning is given by $\mathcal{L}_{total} := \lambda_{\fwk{}}\tilde{\mathcal{L}}_{\fwk{}} + \lambda_{reg} \mathcal{L}_{reg}$.
For all fine-tuned models, the image encoder is frozen and we only update weights of the text encoder, training for a single epoch on the \dataset train data. Following the pretraining procedure of the models under consideration, we unit-normalize all embeddings before calculating losses (i.e. we constrain embeddings to the surface of the unit sphere). Loss calculation details and training hyperparameters are detailed in the supplementary material.

\subsection{Test Datasets and Metrics} \label{sec:exp_test}

We evaluate all models both zero-shot and after alignment (fine-tuning) on the \dataset{} train set. We propose evaluation metrics on \dataset{} to quantify multimodal hierarchical understanding, contrasting with the purely qualitative evaluations provided in prior works~\cite{vendrov2015order, desai2023hyperbolic}. We complement this by evaluating on various external datasets measuring hierarchical and multimodal understanding in other settings. We describe our evaluation procedure below.

\smallskip
\noindent \textbf{\dataset{}}. We perform hierarchical image--text matching on \dataset{} using the methodology of Desai et al.~\cite{desai2023hyperbolic}  by selecting 50 equally-spaced points between the root node and the closest text embedding to the given image; at each point we retrieve the text closest to the image (via cosine similarity) within the given radius of the image. We calculate standard retrieval metrics (precision and recall) relative to the four ground-truth captions for each image. In addition, we calculate the order-aware metric $\tau_d$ to check that ground-truth texts are correctly ordered by their distances from the root node (i.e. the embedding of the empty string); for a given image, this equals the Kendall correlation between $d_\mathbf{r}(\mathbf{e})$ for each ground-truth text's embedding $\mathbf{e}$, and their ground-truth order. In particular, this equals $1$ if and only if all ground-truth texts are ordered correctly by these distances. Note that unlike the retrieval metrics previously described, this metric measures to what extent the textual embedding space agrees with the expected (ground truth) hierarchical structure. These are calculated relative to the manually-curated test set; for qualitative results we perform retrieval over an expanded set of candidate texts (see supp.) to yield more extensive hierarchies.

\smallskip
\noindent \textbf{HyperLex}. The HyperLex dataset~\cite{vulic2017hyperlex} is a benchmark for lexical entailment understanding. We use the standard metric of Spearman correlation with ground-truth entailment scores, using the exterior angle $\xipos$ between embeddings of word pairs as the predicted entailment score. As in prior works, we report this score on all items ($\rho_{all}$) as well as restricted to nouns ($\rho_N$). See the supplementary material for details on prompts used.

\smallskip
\noindent \textbf{BREEDS}. The BREEDS dataset~\cite{santurkar2020breeds} contains a subset of images from ImageNet annotated with two-tiered hierarchical labels (e.g. \literal{stringed instrument} $\to$ \literal{banjo}); we use the subsets\footnote{living17, nonliving26, entity13, entity30} of BREEDS selected by Novack et al.~\cite{novack2023chils} and test the ability of our models to predict both the coarse and fine-grained labels for a given image (in the correct order) out of all labels of all granularities. In order to predict ordered pairs of labels, we consider all pairs of distinct labels (x, y) where the embedding of x is closer to the entailment root, and calculate the score
$c_x + c_y + C \cdot \xir(x, y)$,
where $c_x$ is the CLIP similarity between $x$ and the image (and similarly $c_y$), $C$ is a constant chosen by cross-validation, and $\xir(x, y)$ is the exterior angle between the embeddings of texts $x$ and $y$. We predict the highest scoring $k \in \{1, 5\}$ pairs of labels and report recall at $k$.

\smallskip
\noindent \textbf{Standard multimodal benchmarks}. We evaluate on various standard cross-modal tasks to test whether pretraining knowledge is preserved. Results for cross-modal retrieval on MS-COCO~\cite{lin2014microsoft,chen2015microsoft} and zero-shot image classification on CIFAR-10 and CIFAR-100~\cite{Krizhevsky09learningmultiple} are given here; results on additional datasets are given in the supplementary material.

\begin{table}[t]
  \centering
  \setlength{\tabcolsep}{2pt}
  \begin{tabularx}{0.94\columnwidth}{lcccccccccccccccc}
    \toprule
    &
    \multicolumn{3}{c}{\dataset{}} & &
    \multicolumn{2}{c}{HyperLex} & &
    \multicolumn{2}{c}{BREEDS} & &
    \multicolumn{2}{c}{COCO} & &
    C10 & C100
    \\
    \cmidrule(lr){2-4}
    \cmidrule(lr){6-7}
    \cmidrule(lr){9-10}
    \cmidrule(lr){12-13}
     \cmidrule(lr){15-15}
     \cmidrule(lr){16-16}
    Model &
    P & R & $\tau_d$
    & & $\rho_{all}$ & $\rho_N$
    & & R@1 & R@5
    & & R@1 & R@5
    & & acc & acc
    \\
    \midrule
    $\text{CLIP}^{\text{B}}$ & 0.14 & 0.36 & 0.89 &  & 0.44 & 0.48 &  & 0.22 & 0.50 & & 0.30 & 0.55
    & & \underline{0.90} & \textbf{0.66}
    \\
    $\text{CLIP}^{\text{B}}_{\text{FT}}$ & \textbf{0.15} & \textbf{0.47} & \textbf{0.99} &  & \textbf{0.51} & \textbf{0.55} &  & \textbf{0.24} & \textbf{0.55} & & \textbf{0.31} & \textbf{0.56}
    & & \underline{0.90} & 0.65
    \\
    \midrule
    $\text{CLIP}^{\text{L}}$ & \textbf{0.16} & 0.37 & 0.88 &  & 0.42 & 0.44 &  & 0.29 & 0.61 & & \underline{0.36} & 0.60
    & & \underline{0.96} & 0.78
    \\
    $\text{CLIP}^{\text{L}}_{\text{FT}}$ & 0.15 & \textbf{0.44} & \textbf{0.97} &  & \textbf{0.50} & \textbf{0.54} &  & \textbf{0.32} & \textbf{0.65} & & \underline{0.36} & \textbf{0.61}
    & & \underline{0.96} & \textbf{0.79}
    \\
    \midrule
    $\text{OpenCLIP}^{\text{B}}$ & \textbf{0.16} & 0.33 & 0.87 &  & 0.34 & 0.37 &  & 0.23 & 0.50 & & \underline{0.39} & \underline{0.65}
    & & \textbf{0.94} & 0.76
    \\
    $\text{OpenCLIP}^{\text{B}}_{\text{FT}}$ & 0.14 & \textbf{0.40} & \textbf{0.98} &  & \textbf{0.49} & \textbf{0.55} &  & \textbf{0.25} & \textbf{0.56} & & \underline{0.39} & \underline{0.65}
    & & 0.93 & \textbf{0.77}
    \\
    \midrule
    $\text{OpenCLIP}^{\text{H}}$ & \underline{0.16} & 0.32 & 0.83 &  & 0.06 & 0.03 &  & 0.23 & 0.50 & & \underline{0.48} & \underline{0.73}
    & & \underline{0.98} & \underline{0.86}
    \\
    $\text{OpenCLIP}^{\text{H}}_{\text{FT}}$ & \underline{0.16} & \textbf{0.36} & \textbf{0.97} &  & \textbf{0.37} & \textbf{0.39} &  & \textbf{0.31} & \textbf{0.65} & & \underline{0.48} & \underline{0.73}
    & & \underline{0.98} & \underline{0.86}
    \\
    \midrule
    ALIGN & \underline{0.16} & 0.36 & 0.89 &  & 0.35 & 0.37 &  & 0.21 & 0.52 & & \underline{0.42} & \underline{0.67}
    & & \underline{0.78} & \underline{0.53}
    \\
    ALIGN\textsubscript{FT} & \underline{0.16} & \textbf{0.42} & \textbf{0.96} &  & \textbf{0.44} & \textbf{0.47} &  & \textbf{0.23} & \textbf{0.58} & & \underline{0.42} & \underline{0.67}
    & & \underline{0.78} & \underline{0.53}
    \\
    \bottomrule
  \end{tabularx}
  \caption{\textbf{Results for hierarchical text--image matching on \dataset{} (test set), existing hierarchical understanding benchmarks, and standard text$\to$image retrieval on COCO (val set)}. Best results are in \textbf{bold}; ties are \underline{underlined}. Superscript letters indicate model size. C10 and C100 indicate CIFAR-10 and 100 respectively. P and R are precision and recall, and $\tau_d$ is the order-aware metric defined in Section \ref{sec:exp_test}. $\rho_{all}$ and $\rho_N$ are Spearman correlation between predicted and ground-truth values for all items and for nouns. R@k refers to recall at k and acc refers to categorical accuracy. As seen above, fine-tuning mostly improves hierarchical understanding without significantly impacting standard cross-modal task performance.}
\label{tab:results}
\end{table}

\begin{figure*}
    \centering

        \begin{minipage}[t]{0.9\linewidth}
        \centering
        \includegraphics[width=0.2\linewidth]{figures/qualitative/snail.jpg}
        \hspace{85pt}
        \includegraphics[width=0.2\linewidth]{figures/qualitative/cats.jpg}
    \end{minipage}
    
    \begin{minipage}[t]{0.95\linewidth}
        \begin{tabularx}{\linewidth}[t]{CC|CC}%
            \multicolumn{1}{c}{\scriptsize{\textbf{CLIP}}} & \multicolumn{1}{c}{\scriptsize{\textbf{+alignment}}} & \multicolumn{1}{c}{\scriptsize{\textbf{CLIP}}} & \multicolumn{1}{c}{\scriptsize{\textbf{+alignment}}}
            \\
            \rowcolor{Orange!10} \qual{fun} & \qual{food animal}
            &\cellcolor{Cyan!10} \qual{two} & \cellcolor{Cyan!10} \qual{cat}
            \\
            \rowcolor{Orange!20} \qual{top} & \qual{vegetables} 
            &\cellcolor{Cyan!20} \qual{sleep} & \cellcolor{Cyan!20} \qual{cats}
            \\
            \rowcolor{Orange!30} \qual{...} & \qual{...} 
            &\cellcolor{Cyan!30} \qual{...} & \cellcolor{Cyan!30} \qual{...}
            \\
            \rowcolor{Orange!40} \qual{a close up of a plate of food with broccoli} & \qual{A raw piece of broccoli with something growing from it.}
            &\cellcolor{Cyan!40} \qquad\newline\qual{cats sleeping with a remote} & \cellcolor{Cyan!40} \qual{Couple of cats sleeping on opposite ends of the couch}
            \\
            \rowcolor{Orange!50} \qual{A worm sits on top of a piece of broccoli.} & \qual{A worm sits on top of a piece of broccoli.}
            &\cellcolor{Cyan!50} \qual{Two cats sleeping with a remote control near each of them.} & \cellcolor{Cyan!50} \qual{Two cats sleeping with a remote control near each of them.}
            \\
        \end{tabularx}
    \end{minipage}

    \caption{\textbf{Qualitative hierarchical text-image matching results on \dataset{}} (test set). The left column of each table shows hierarchical text-image matching applied using pretrained CLIP-Large, while the right column shows results on the same images after alignment (fine-tuning). The results above are abridged; for full results, see the supplementary material.}
    \label{fig:qualitative}
\end{figure*}

\subsection{Results and Discussion} \label{sec:exp_res}

Quantitative results are displayed in Table \ref{tab:results}. Across tasks and datasets, we see that our \fwk{} framework demonstrates that models like CLIP possess hierarchical understanding in the zero-shot regime, while our fine-tuning generally further enhances performance across model types and sizes. In addition, Figure \ref{fig:qualitative} shows qualitative results of hierarchical image--text matching with CLIP on \dataset{}. We see that CLIP already shows the emergent ability to perform hierarchical retrieval, although its quality is further improved by model alignment. We note that this improvement is most evident for more general terms, consistent with prior work observing that foundation VLMs may struggle with matching highly general terms to images~\cite{novack2023chils,xu2023challenges}. In the supplementary material, we further show that this improvement is seen even when controlling for text length.

As seen in the cross-modal retrieval results for COCO in Table \ref{tab:results} (and as shown on additional datasets and tasks in the supplementary material), our fine-tuning has a negligible effect on standard multimodal tasks. This supports our fine-tuning being a type of model alignment, bringing latent knowledge of hierarchies to the surface without significantly impacting pretrained knowledge.

Additionally, our results on existing hierarchical understanding datasets in Table \ref{tab:results} show that VLMs can be probed and aligned with the \fwk{} framework for more general hierarchical understanding, beyond hierarchical image--text matching and the specific format of \dataset{}. Results on BREEDS show the applicability of our methodology to other multimodal datasets. Results on HyperLex demonstrate the surprising finding that VLMs such as CLIP with alignment applied perform competitively to methods using dedicated embeddings for the unimodal (text-only) task of lexical entailment prediction. For example, our best $\rho_{all}$ is $0.51$ versus $0.69$ for LEAR~\cite{vulic2017specialising} and $0.59$ for DOA-E~\cite{athiwaratkun2018hierarchical}, both of which were trained explicitly on lexical entailment data; see the supplementary material for an extensive comparison to prior methods. While lexical entailment prediction is a non-obvious task for VLMs as it is purely unimodal (text-only), we hypothesize that paired image-text data naturally provides supervision for learning concept hierarchies (supported by prior works discussed in Section \ref{sec:rw}), and that this knowledge is unlocked by our alignment procedure.

\begin{figure}
    \begin{minipage}{0.1\linewidth}
        \vspace{12pt}
        \includegraphics[trim={6cm 2cm 2cm 3cm},clip,width=0.9\linewidth]{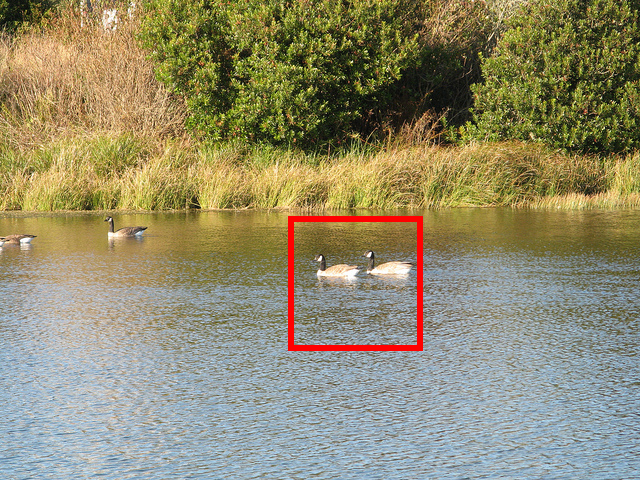}
        \includegraphics[width=0.9\linewidth]{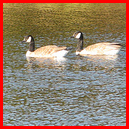}
    \end{minipage}
    \hfill
    \begin{minipage}{0.9\linewidth}
        \begin{tabularx}{\linewidth}[t]{C|C}
            \multicolumn{1}{c}{\scriptsize \textbf{Ours}} & \multicolumn{1}{c}{\scriptsize \textbf{MERU}} \\
            \cellcolor{Dandelion!10} \qual{photo} & \cellcolor{Bittersweet!10}
            \qual{out} \\
            \cellcolor{Dandelion!20} \qual{aquatic animals} & \cellcolor{Bittersweet!20} \qual{members} \\
            \cellcolor{Dandelion!30}
            \vspace{-2pt}
            \vfill
            \qual{waterfowl} & \cellcolor{Bittersweet!30}
            \qual{learning environment} \\
            \cellcolor{Dandelion!40}
            \qual{...}
            & \cellcolor{Bittersweet!40}
            \qual{...} \\
            \cellcolor{Dandelion!50}
            \qual{A gaggle of geese swim in a body of water}
            & \cellcolor{Bittersweet!50}
            \qual{family swans swimming on a lake}
        \end{tabularx}
    \end{minipage}
    \caption{\textbf{Comparison to MERU~\cite{desai2023hyperbolic}.} We show an image from the \dataset{} test set and a zoomed-in view bordered in red, along with our (aligned CLIP-L) hierarchical text-image matching results and those of MERU (ViT-L). As seen above, our fine-tuned model yields more semantically plausible results with a clear hierarchical structure.}
    \label{fig:meru}
\end{figure}

We also provide a comparison to MERU~\cite{desai2023hyperbolic}, a recent hierarchical understanding model which is trained from scratch (rather than building on top of an existing pretrained foundation model). We use the strongest MERU variant, a hyperbolically-trained dual encoder model with a ViT-L backbone, and compare our \fwk{} results on those of the comparable CLIP-L model (reported in Table \ref{tab:results}). For standard multimodal metrics, we reproduce MERU's reported $0.32$ R@5 on COCO text$\to$image (vs. our $0.61$ for fine-tuned CLIP-L). We also perform hierarchical retrieval on \dataset{} (using the hyperbolic inference procedure of Desai et al. to match MERU's learned geometry), yielding precision 0.11, recall 0.11, and $\tau_d$ 0.79, far underperforming our results.
As seen in Figure \ref{fig:meru}, these metrics reflect weaker performance on both semantic text-image matching and hierarchical knowledge, emphasizing the importance of strong foundation VLMs for downstream tasks such as ours.

\subsection{Ablations} \label{sec:ablations}

To justify the effectiveness of our framework relative to the existing EC framework (described in Section \ref{sec:prelim}), we align CLIP-Base while replacing the loss $\mathcal{L}_{\fwk}$ in fine-tuning with an EC loss. Following Ganea et al.~\cite{ganea2018hyperbolic}, we apply this to both positive and negative examples by using the loss $\mathcal{L}_{EC} = \left[ \pm(\xipos - \thre) \right]^+$, where the inner sign is negative for negative examples. Here the half-aperture angle defined by $\sin \thre = \min(1, 0.05 / \dre)$ For hierarchical image--text matching on \dataset{}, this yields precision 0.13 and recall 0.30, significantly underperforming our \fwk{} alignment results as seen in Table \ref{tab:results}.

In the supplementary material, we further ablate each key element of our framework to show their necessity, including the use of pretrained weights followed by fine-tuning with our loss; the use of a learnable root initialized with the empty string embedding (rather than the fixed roots used in prior works~\cite{ganea2018hyperbolic, desai2023hyperbolic}); the use of regularization loss, and hard negative mining; and the structure of our training data (four-tiered, with positive and negative items). We also show that learning the position of the root vector alone (while keeping other embeddings fixed) is insufficient, and we compare to additional EC framework variants.
\section{Conclusion}
\label{sec:conclusion}

In our work, we study the emergent capability of foundation VLMs to understand visual-semantic hierarchies, proposing the \framework{} (\fwk{}) framework to probe and optimize them for hierarchical understanding. We present the \dataset{} dataset and benchmark for hierarchical text-image matching, along with evaluation metrics for this task. We show that these models demonstrate emergent hierarchical understanding, outperforming prior methods designed explicitly for representing hierarchical structures even when used zero-shot, with an additional overall performance boost after fine-tuning.

\newcommand{\myurl}{https://commons.wikimedia.org/wiki/File:Carica_papaya_-_K\%C3\%B6hler\%E2\%80\%93s_Medizinal-Pflanzen-029.jpg}

Regarding limitations, we focus on the representation learning setting using dual encoder VLMs, while other architectures might require a different analysis.
Additionally, while we assign a single linear hierarchy of texts to an image, the semantic hierarchy of texts describing an image naturally form a branching structure, as they may focus on different visual details.
Future work could explicitly model this branching structure to automate organization of image collections similarly to the manually-organized branching hierarchical structure of widely-used online image repositories such as Wikimedia Commons.

\section*{Acknowledgements}

We thank Yotam Elor, Roi Livni, Guy Tevet, Chen Dudai, and Rinon Gal for providing helpful feedback.
This work was partially supported by ISF (grant number 2510/23).

\bibliographystyle{splncs04}
\bibliography{main}

\clearpage
\appendix
{\LARGE\textbf{Supplementary Material}}

\section{Interactive Visualization Tool} \label{sec:supp_viz}

Please see \href{https://hierarcaps.github.io/web/viz.html}{our project page} for an interactive visualization of model results on the \dataset{} test set. There we also provide visualizations of a random subset of the \dataset{} train set.

\section{Additional Details}

\subsection{\dataset{} Construction}

Below we provide a full technical description of the procedure used to construct \dataset{}. Examples of items from the train set of \dataset{}, along with the full test set, may be viewed in our interactive visualization tool.

\subsubsection{Models and Inference Settings (\dataset{})} \label{sec:supp_data_models}

In this section, we describe the models used in producing \dataset{} and their inference settings.

To generate text for \dataset{}, we use the LLM Llama-2~\cite{touvron2023llama} to produce hierarchies from existing captions in a zero-shot manner. We use the following model checkpoints (from the Hugging Face model hub):

\begin{itemize}
\item \texttt{meta-llama/Llama-2-13b-chat-hf} (henceforth $\text{LLM}_{chat}$)
\item \texttt{meta-llama/Llama-2-13b-hf} (henceforce $\text{LLM}_{base}$)
\end{itemize}

We use the former instruction-tuned model for zero-shot tasks including instruction prompts (Section \ref{sec:supp_data_seeds}). We use the latter for few-shot augmentation of new synthetic hierarchies (Section \ref{sec:supp_data_fewshot}). We generate text using stochastic sampling with temperature (1.2 for $\text{LLM}_{chat}$ and 1.0 for $\text{LLM}_{base}$). In all cases, we load models using 4-bit quantization as implemented \texttt{bitsandbytes} integration with the Hugging Face \texttt{transformers} library, to enable inference on a single NVIDIA A5000 GPU. 

In addition to generation, we also use a pretrained Natural Language Inference (NLI) model to filter for the desired logical properties (entailment or contradiction). In particular, we use BART-Large~\cite{lewis2019bart} fine-tuned on the MNLI dataset~\cite{mnli}, using the checkpoint \texttt{facebook/bart-large-mnli} from the Hugging Face model hub. When providing text to the NLI model, we use the prompt \texttt{a picture of: "\{\}"}, inserting the text into the slot indicated by \texttt{\{\}}.

For the smaller model $\mathcal{H}$ used in our knowledge distillation procedure, we use the encoder-decoder model Flan-T5~\cite{chung2022scaling}, using the \texttt{google/flan-t5-base} checkpoint from the Hugging Face model hub. Training details for this model are described in Section \ref{sec:supp_data_kd}. When generating text with this model, we use stochastic sampling with temperature 1.0. 

\subsubsection{Seed Data Generation} \label{sec:supp_data_seeds}

We begin by producing a small (${\sim}900$-item) set of caption hierarchies. This is motivated by the fact that we can produce these using real captions with a time-consuming and failure-prone process, but having these seeds will later allow us to produce hierarchies more reliably (see Sections \ref{sec:supp_data_kd}--\ref{sec:supp_data_final}).

To produce four-tiered caption hierarchies, we first select captions from Conceptual Captions~\cite{sharma2018conceptual} (CC) containing at least ten words, placing these in the fourth tier ($T_4$). We then apply prompts from Table \ref{tab:supp_prompts} using $\text{LLM}_{chat}$ (see Section \ref{sec:supp_data_models}) in order to produce preceding tiers; in other words, we apply the first prompt with the text of $T_4$ inserted in the blank slot ({\tt \{\}}) to produce $T_3$, we then insert $T_3$ into the blank slot of the next prompt to produce $T_2$, and similarly we insert $T_2$ into the last prompt to produce $T_1$. Note that, in addition to the instruction delimiters ({\tt \myinst \mysinst}), the prompts contain partial answers by design, as this enforces a uniform format for answers. LLM outputs are postprocessed by selecting only for text enclosed in quotation marks to avoid irrelevant text in outputs, and by removing texts that do not contain standard alphanumeric characters.

To ensure that these hierarchies have the desired logical structure, we filter each generated tier using the pretrained NLI model (described in Section \ref{sec:supp_data_models}). In particular, to each pair of tiers $(T_i, T_j)$ with $i < j \; (i, j \in \{1, 2, 3, 4\})$ we apply the NLI model to output probabilities $(p^{(ij)}_e, p^{(ji)}_e)$, where $p^{(ij)}_e$ is the probability that $T_i$ logically entails $T_j$, and $p^{(ji)}_e$ is the probability that $T_j$ logically entails $T_i$. We filter so that for $i < j$, we have $p^{(ij)}_e < 0.5$ and $p^{(ji)}_e > 0.5$. In other words, we ensure that (up to the predictions of the NLI model) each tier is logically entailed by all following tiers, but does not entail them itself. Hence, each tier should be strictly more general than the following tiers.

Due to time and compute constraints, we run this process on a 5K-item subset of CC, yielding ${\sim}900$ seed hierarchies. Examples of such data include the following (displayed in the format $T_1 \to T_2 \to T_3 \to T_4$):
\begin{itemize}
\item \emph{cutting tools $\to$ scissors $\to$ middle-aged woman cutting white fabric with scissors in super slow motion $\to$ super slow motion of middle-age woman hands with scissors cutting white fabric on a tabletop}
\item \emph{dessert $\to$ pie $\to$ decorated pie on a white background $\to$ pie with candles and gifts in boxes on a white background}
\item \emph{youth $\to$ boys $\to$ boys setting up tent by lake $\to$ teenage boys pitching a tent at the lake in forest}
\end{itemize}

\subsubsection{Few-Shot Synthetic Data Augmentation} \label{sec:supp_data_fewshot}

Our seed data, described in the previous section, has the correct format but is relatively small. To obtain sufficient data for knowledge distillation, we augment by generating synthetic caption hierarchies with the same format. In particular, we provide seed hierarchies as few-shot examples to $LLM_{base}$ (see Section \ref{sec:supp_data_models}) and then generate a new line of text. As $LLM_{base}$ is not instruction-tuned, no prompt text is needed. At each iteration, we provide ten randomly-selected examples of seed hierarchies as input to $LLM_{base}$, separated by newlines, and we proceed to generate text until another newline is predicted. We filter for outputs that have exactly four tiers. Note that we do not filter for logical entailment, as we find that the outputs sufficiently reflect the desired logical structure for the purposes of knowledge distillation (described in the next section) when used as-is.

While this process can be run indefinitely, we terminate this when we have generated ${\sim}10K$ synthetic caption hierarchies. Examples of generated synthetic data include the following (displayed in the format $T_1 \to T_2 \to T_3 \to T_4$):
\begin{itemize}
\item \emph{music artist $\to$ singer $\to$ singer with group performs at event $\to$ pop singer with group performs at event}
\item \emph{textiles $\to$ towels $\to$ striped towels $\to$ a variety of striped towels in a range of colours and patterns - for sale on amazon}
\item \emph{flowers $\to$ poppies $\to$ fields of poppies. $\to$ a field of poppies, bright orange, on a sunny day}
\end{itemize}

\subsubsection{Hierarchy Completion Distillation} \label{sec:supp_data_kd}

In order to learn to efficiently complete hierarchies from existing captions, we train a small language model $\mathcal{H}$ using the ${\sim}10K$ synthetic hierarchies from Section \ref{sec:supp_data_fewshot} as supervision, thus distilling their contents into $\mathcal{H}$. We select the encoder-decoder model Flan-T5~\cite{chung2022scaling}, described further in Section \ref{sec:supp_data_models}, and define its input and target texts as follows: 

Ground-truth hierarchies are represented as a single string $T_1 \to T_2 \to T_3 \to T_4$ using a special token $\to$ to delimit tiers. At each step, we randomly select up to three initial tiers as the encoder prompt, and use the remaining tiers as targets. In addition, we randomly (with probability $0.5$) reverse the directionality and use the special token $\leftarrow$ instead (i.e. $T_4 \leftarrow T_3 \leftarrow T_2 \leftarrow T_1$).  Hence, possible input-target pairs include:

\begin{itemize}
\item Input: $T_1 \to$; Target: $T_2 \to T_3 \to T_4$
\item Input: $T_1 \to T_2 \to$; Target: $T_3 \to T_4$
\item Input: $T_4 \leftarrow T_3 \leftarrow$; Target: $T_2 \leftarrow T_1$
\end{itemize}

Note that this includes the case where no initial tiers are selected; we use the delimiter $\to$ or $\leftarrow$ as the input so the model has an indication as to the desired directionality of the output. In other words, these input-target pairs are:

\begin{itemize}
\item Input: $\to$; Target: $T_1 \to T_2 \to T_3 \to T_4$
\item Input: $\leftarrow$; Target: $T_4 \leftarrow T_3 \leftarrow T_2 \leftarrow T_1$
\end{itemize}

We train $\mathcal{H}$ on such inputs-target pairs using a language modelling objective. We train for three epochs on our synthetic hierarchies, using a batch size of 32, learning rate 1e-3, and the AdamW optimizer.$\mathcal{H}$ thus learns to complete hierarchies in either direction, and inference with $\mathcal{H}$ is faster and more reliable than the zero-shot LLM inference procedure described in Section \ref{sec:supp_data_seeds}.

\subsubsection{Final Data Generation (\dataset{})} \label{sec:supp_data_final}

After training $\mathcal{H}$ to complete hierarchies, as described in the previous section, we apply it to captions from the Conceptual Captions~\cite{sharma2018conceptual} (CC) and MS-COCO~\cite{lin2014microsoft,chen2015microsoft} datasets in order to put them in the context of full caption hierarchies for \dataset{}. We use captions from the CC train set to produce the \dataset{} train set, and captions from the COCO validation set to produce the \dataset{} test set, selecting captions with at least ten words.

We proceed as follows: Given a ground-truth caption $C = T_4$, we provide $T_4 \leftarrow$ as input to $\mathcal{H}$ and generate a batch of $64$ output candidates. We filter these candidates for those of the format $T_3 \leftarrow T_2 \leftarrow T_1$ (i.e. exactly three output tiers), and then check if any of these candidates obey the following requirements: None of the $T_i$ may be empty, no two adjacent tiers are identical (i.e. $T_1 \neq T_2 \neq T_3 \neq T_4$), $T_3$ must be at least two words shorter than $T_4$, $T_3$ must be at least ten characters shorter than $T_4$. Additional, for $i \in \{0, 1, 2\}$, we require that $T_i$ be entailed by $T_{i+1}$ while not entailing it (using our pretrained NLI model and following the procedure of Section \ref{sec:supp_data_seeds}). We output the first candidate obeying these conditions, if any satisfy them.

For \dataset{} test set items, we use the above procedure alone to generate hierarchies of positive texts (i.e. texts that are consistent with the ground-truth caption). For train set items, we additionally add negative examples (i.e. texts that contradict the ground-truth caption) as follows: Given the positive hierarchy $T_1 \to T_2 \to T_3 \to T_4$, where $T_4$ is the original caption, we now apply $\mathcal{H}$ in the opposite direction to deliberately hallucinate negative texts. In particular, we provide the following prompts to $\mathcal{H}$:

\begin{itemize}
\item $\to$
\item $T_1 \to$
\item $T_1 \to T_2 \to$
\item $T_1 \to T_3 \to T_3 \to$
\end{itemize}

We then generate text with $\mathcal{H}$ until the next $\to$ token, or until the end-of-string token is output (whichever comes first). $\mathcal{H}$ proceeds to generate outputs $T'_1, T'_2, T'_3, T'_4$ respectively. For each such $T'_i$, we input the pair $(T_i, T'_i)$ into our pretrained NLI model (described in Section \ref{sec:supp_data_models}) and calculate $p^{(i)}_c$, the probability that these texts contradict each other; we filter to require that $p^{(ii')}_c > 0.5$. Additionally, if $i > 1$, we input the pairs $(T_{i-1}, T'_i)$ and $(T'_i, T_{i-1})$ into the NLI model and check that $T_{i-1}$ is entailed by $T'_i$ but does not entail it, following the procedure of Section \ref{sec:supp_data_seeds}. As this procedure is failure-prone, we repeat it eight times (saving any valid $T'_i$ values that are produced). If we finish with $T'_1, T'_2, T'_3, T'_4$ values that obey the previous conditions, we output them as the desired negative examples.

This inference procedure yields $73K$ train set hierarchies, containing positive and negative texts, based on captions from CC. For our test set, after applying inference to captions from COCO to obtain positive hierarchies, we manually filter and correct a subset to obtain a hand-validated $1K$-item test set as a clean quantitative evaluation benchmark. While metrics are calculated using these manually-curated items alone, for qualitative results, we perform retrieval using the expanded set of candidate texts consisting of all automatic generations before hand-filtering, to yield more extensive predicted hierarchies. In our interactive visualization tool, we provide a visualization of all \dataset{} test set items along with a random subset of \dataset{} train set items.

\begin{figure*} { \tt \scriptsize \begin{center}
\begin{tabular}{l}
\toprule
\\
\myinst \\
Shorten the following image caption. Feel free to remove details. \\
Put your answer in quotation marks: " " \\
\\
"\{\}" \\
\mysinst \\
\\
Sure! Here is the original caption and its shortened version: \\
\\
Original caption: "\{\}" \\
Shortened version: \\
\\
\midrule
\\
\myinst \\
You are given the following caption to an image: \\
\\
"\{\}" \\
\\
What one or two words should be filled in the blank in the following \\
alternative caption for the same image? \\
\\
"A picture of (a/the) \_\_\_ ." \\
\\
\mysinst \\
\\
The best word or words to fill in the blank would be: "A picture of (a/the) \\
\\
\midrule
\\
\myinst \\
What term describes a general concept that includes "\{\}"? \\
Answer in one or two words, and put your answer in quotation marks: " " \\
\mysinst \\
\\
The best term that includes the given word or phrase is "  \\
\\
\bottomrule
\end{tabular}\end{center} } 
\caption{Prompts used in the construction of \dataset{}. {\tt \{\}} indicates the slot for inserting an input text. {\tt \myinst} and {\tt \mysinst} are special tokens used by the LLM Llama-2 to delineate the given instruction. Note that the prompts contain partial answer text by design, in order to enforce a uniform output format that may be easily parsed.}  
 \label{tab:supp_prompts}  
\end{figure*}
\subsection{Entailment Cone Details}

We provide further details related to the Entailment Cone (EC) framework of Ganea et al.~\cite{ganea2018hyperbolic} and its relation to our framework.

\subsubsection{Partial Order Relations}

As discussed in the main paper, the EC framework defines a \emph{partial order} on embeddings. We proceed to provide the mathematical definition of this term and its significance to entailment hierarchies.

A partial order on a set $X$ is a binary relation $\leq$ defined on $X \times X$ which is reflexive ($x \leq x \;\forall\; x \in X$), antisymmetric ($a \leq b$ and $b \leq a$ imply $a = b$) and transitive ($a \leq b$ and $b \leq c$ imply $a \leq c$). Elements $a, b \in X$ are called \emph{comparable} if either $a \leq b$ or $b \leq a$. $X$ along with its associated partial order is called a partially ordered set or \emph{poset}.

Partial orders have a natural application in the theory of logic, as logical entailment naturally forms a partial order over the the semantic denotations of assertions (i.e. over meanings)~\cite{toledo2013semantic}.  In particular, multimodal text-image data can be associated with a poset whose elements are the semantic denotations of images and texts\footnote{It is not a partial order over the images and texts themselves, but rather a \emph{preorder}, as two distinct items can be logically equivalent (e.g. ``this is a big cat'' vs. ``this is a large cat'').}, and whose partial order is identified with visual and textual entailment~\cite{vendrov2015order}. This motivates the design of the EC framework, which induces a partial order on embeddings corresponding to a logical entailment relation between items.

\subsubsection{Derivation of \fwk{} Loss from Margin EC Loss} \label{sec:supp_ec_deriv}

We proceed to show that our \fwk{} loss function can be derived as the limit of a margin EC loss applied to contrastive pairs as the margin tends to infinity. See Section \ref{sec:supp_ec_comparison} for a comparison of results between our framework and training with an EC-based loss with various margins.

Consider the EC margin loss
$$\mathcal{L}_{\alpha} := \max(-\alpha, \pm (\xipos - \thetad)),$$
where the sign ($\pm$) is $1$ for positive entailment pairs and $-1$ for negative pairs (i.e. pairs $\mathbf{e}, \mathbf{e}'$ where $\mathbf{e}$ does not entail $\mathbf{e}'$). We find (in Section \ref{sec:supp_ec_comparison}) that performance improves for larger margins, and the limiting case ($\alpha \to \infty$) is
$$\mathcal{L}_{\infty} := \pm (\xipos - \thetad).$$

We now use our key insight that the sum of such losses simplifies dramatically when for each positive pair $\mathbf{e}, \mathbf{e}'$ there is a negative pair $\mathbf{e}, \mathbf{e}''$. This matches the design of \dataset{}; for example, $\mathbf{e}, \mathbf{e}', \mathbf{e}''$ could be the embeddings of the texts $t=$\litquot{animal}, $t'=$\litquot{goat}, and $t''=$\litquot{portrait} respectively, as $t$ is entailed by $t'$ and $t$ contradicts $t''$ (and thus is not entailed by $t''$). Summing losses on these pairs gives
\begin{align*}
& \mathcal{L}_\infty(\mathbf{e}, \mathbf{e}') + \mathcal{L}_\infty(\mathbf{e}, \mathbf{e}'') \\
= & \: \xipos - \thetad  - \xineg + \thetad \\
= & \: \xipos - \xineg,
\end{align*}
which motivates the definition of the \fwk{} loss as
$$\ourloss := \xipos - \xineg.$$

Note that this loss does not depend on the half-aperture function $\theta$.
\subsection{Model Details}

\begin{table}[t]
  \centering
  \begin{tabularx}{0.5\columnwidth}{lcc}
    \toprule
    \textbf{Model} & \textbf{\#Param} & \textbf{\#Param (text)} \\
    \midrule
    CLIP$^\text{B}$ & 151M & 63M \\
    CLIP$^\text{L}$ & 428M & 123M \\
    OpenCLIP$^\text{B}$ & 151M & 63M \\
    OpenCLIP$^\text{H}$ & 986M & 353M \\
    ALIGN & 172M & 109M \\
    \bottomrule
  \end{tabularx}
  \caption{Parameter counts of models used. Superscript letters indicate model size. \#Param indicates total number of parameters, and \#Param (text) indicates the number of parameters in the model's text encoder alone.}
\label{tab:supp_models}
\end{table}

For our results on hierarchical understanding tasks, we use the following model checkpoints (all from Hugging Face model hub):

\begin{itemize}
    \item \texttt{openai/clip-vit-base-patch32}
    \item \texttt{openai/clip-vit-large-patch14}
    \item \texttt{laion/CLIP-ViT-B-32-laion2B-s34B-b79K}
    \item \texttt{laion/CLIP-ViT-H-14-laion2B-s32B-b79K}
    \item \texttt{kakaobrain/align-base}
\end{itemize}

These correspond respectively to the models CLIP-Base, CLIP-Large, OpenCLIP-Base, OpenCLIP-Huge, and ALIGN respectively.  See Table \ref{tab:supp_models} for parameter counts of these models.
\subsection{Training Details} \label{sec:suppl_hyper}

When performing model fine-tuning, we train for a single epoch on \dataset{}, using batch size 8 and the AdamW optimizer with learning rate 1e-7. For each model under consideration, we only train its text encoder while keeping all other weights frozen.

The total loss used for training is $\mathcal{L}_{total} = \lambda_{\fwk{}}\tilde{\mathcal{L}}_{\fwk} + \lambda_{reg}\mathcal{L}_{reg}$, as described in the main paper. We use coefficients $\lambda_{\fwk{}} = 1.0$ and $\lambda_{reg} = 10.0$ (ablated in Section \ref{sec:suppl_abl}).

The aggregate \fwk{} loss $\tilde{\mathcal{L}}_{\fwk}$ is calculated as follows: Each item in the \dataset{} train set contains a positive caption hierarchy $(P_1, P_2, P_3, P_4)$ with accompanying negative captions $(N_1, N_2, N_3, N_4)$. As described in the main paper, for each $i \in \{1, 2, 3\}$ the caption triplet $(P_i, P_{i+1}, N_i)$ can be used to calculate a exterior angle-based contrastive loss. For tier $i$ in item $j$ of a minibatch, denote this loss by $\mathcal{L}_{\fwk}^{(i,j)} = \xipos - \xineg$, where $(\mathbf{e}, \mathbf{e}', \mathbf{e}'')$ are the embeddings of the respective elements of the corresponding caption triplet. We take the mean value of this loss across such $i$ and $j$ to yield mean \fwk{} loss $\mathcal{L}^{mean}_{\fwk}$. In addition, we perform hard example mining as follows: Let $(i', j')$ denote the $(i, j)$ values that maximize $\xipos$, and let $(i'', j'')$ denote the $(i, j)$ values that minimize $\xineg$. In other words, these are the indices of items with maximal terms in the \fwk{} loss function. We then define the mini-max loss $\mathcal{L}^{mm}_{\fwk} := \mathcal{L}_{\fwk}^{(i',j')} + \mathcal{L}_{\fwk}^{(i'',j'')}$. Finally we define the aggregate loss $\tilde{\mathcal{L}}_{\fwk} := \mathcal{L}^{mean}_{\fwk} + \mathcal{L}^{mm}_{\fwk}$. This loss both encourages the \fwk{} loss to decrease across all items (first term) as well as focusing on the hardest examples (second term).

\subsection{\dataset{} Inference Details}

For qualitative and quantitative results on the \dataset{} test set, we perform hierarchical image$\to$text retrieval. Following the methodology of Desai et al.~\cite{desai2023hyperbolic}, we select 50 equally-spaced points between the root node and closest text embedding to the given image; at each point, we retrieve the closest candidate text within the given radius, as measured by embedding cosine similarity. As the closest points to the origin often lead to a degenerate solution where only a single text embedding exists within the given radius, we omit the closest prediction to the origin.

Quantitative metrics are calculated with respect to the manually-curated test set of 1K ground-truth caption hierarchies. For qualitative results we use an expanded candidate set of all automatic generations before hand-filtering, to yield more extensive predicted hierarchies.  These consist of text produced using automatic hierarchy completion (see Section \ref{sec:supp_data_final}) on ground-truth reference captions corresponding to each image in the COCO validation set. We release both the manually-curated test set used for metric calculations as well as the expanded set used for qualitative results.

\subsection{External Benchmark Experimental Details}

For tests on HyperLex, we evaluate by inserting lexical items (e.g. \emph{girl}) into the prompt \emph{a picture of a \underline{$\qquad$}} (e.g. \emph{a picture of a girl}) before embedding them, as we observe that this empirically outperforms embedding the lexical items as-is on the HyperLex evaluation metrics. We note the similarity to the common practice of inserting class names into similar prompts when using CLIP for zero-shot classification~\cite{radford2021learning}.

For tests on BREEDS, we reiterate that we predict label pairs $(x, y)$ for an image to maximize the score $c_x + c_y + C \cdot \Xi_{\mathbf{r}}(x, y)$, where $c_x$ and $c_y$ are embedding similarities between each label and the image, $C$ is a constant chosen by cross-validation, and $\Xi_{\mathbf{r}}(x, y)$ is the exterior angle between embeddings of texts $x$ and $y$. Label embeddings are produced using the prompt ensemble of Radford et al.~\cite{radford2021learning}. We select the constant $C$ as follows: We split test items into two equally-sized folds, $F_0$ and $F_1$. For each metric (R@1, R@5), we find the value of $C$ among 20 equally-spaced values between $0$ and $0.2$ which maximizes the metric on $F_i$ and output its value $v_{i-1}$ on $F_{1-i}$. Finally we report the average value $(v_0 + v_1) / 2$.
\section{Additional Results and Comparisons}

\subsection{Ablation Study} \label{sec:suppl_abl}

\begin{table*}[t]
  \centering
  \begin{tabularx}{0.85\textwidth}{lccccccccccccc}
    \toprule
    &
    \multicolumn{3}{c}{\dataset{}} & &
    \multicolumn{2}{c}{HyperLex} & &
    \multicolumn{2}{c}{BREEDS} & &
    \multicolumn{2}{c}{COCO}
    \\
    \cmidrule(lr){2-4}
    \cmidrule(lr){6-7}
    \cmidrule(lr){9-10}
    \cmidrule(lr){12-13}
    Model &
    P & R & $\tau_d$
    & & $\rho_{all}$ & $\rho_N$
    & & R@1 & R@5
    & & R@1 & R@5
    \\
    \midrule
    $\text{CLIP}^{\text{B}}$ & 0.14 & 0.36 & 0.89 &  & 0.44 & 0.48 &  & 0.22 & 0.50 & & 0.30 & 0.55
    \\
    $\text{CLIP}^{\text{B}}_{\text{FT}}$ & 0.15 & 0.47 & 0.99 &  & 0.51 & 0.55 &  & 0.24 & 0.55 & & 0.31 & 0.56
    \\
    \midrule
    fixed root ($\mathbf{\bar{e}}$) & 0.16  & 0.44  & 0.98  &  & 0.50 & 0.55 &  & 0.24 & 0.54 &  & 0.32 & 0.56\\
    fixed root ($\mathbf{0}$) & 0.21 & 0.19 & 0.54  &  & 0.23 & 0.23 &  & 0.04 & 0.09 &  & 0.31 & 0.55 \\
    fixed root ($\mathbf{e}_\emptyset$) & 0.14  & 0.47  & 0.98  &  & 0.50 & 0.55 &  & 0.24 & 0.54 &  & 0.31 & 0.55 \\
    only learning root &  0.15  & 0.37 & 0.93 & & 0.47 & 0.51 & & 0.24 & 0.53 & & 0.29 & 0.53\\
    $\lambda_{reg}=0$ & 0.15 & 0.45 & 0.99 &  & 0.49 & 0.53 &  & 0.21 & 0.52 &  & 0.26 & 0.49 \\
    3 tiers & 0.17 & 0.50 & 0.93 & & 0.40 & 0.44 & & 0.17 & 0.41 & & 0.30 & 0.54 \\
    no negative examples & 0.16 & 0.42 & 0.99 &  & 0.48 & 0.52 &  & 0.25 & 0.54 &  & 0.28 & 0.52 \\
    FT with CLIP objective & 0.14 & 0.28 & 0.95 & & 0.43 & 0.47 & & 0.22 & 0.49 & & 0.32 & 0.58 \\
    training from scratch & 0.00 & 0.01 & 0.93 & & 0.11 & 0.12 & & 0.00 & 0.00 & & 0.00 & 0.00 \\
    \bottomrule
  \end{tabularx}
  \caption{Ablation study results. We first reproduce the results on our suite of tasks for a CLIP model before and after model alignment (fine-tuning). Below this, we display fine-tuning results when removing key components of our framework. Task names and metrics follow the notation of the results table in the main paper.}
\label{tab:supp_abl}
\end{table*}

We show the contribution of core elements of our fine-tuning framework by ablating each in turn. See Table \ref{tab:supp_abl} for quantitative results of these ablations, which we describe below, recalling that the first three benchmarks listed (\dataset{}, HyperLex, BREEDS) measure hierarchical knowledge while the last (COCO) measures standard multimodal understanding (i.e. pretraining knowledge).

We ablate the effect of a learnable root corresponding to the embedding of the empty string ($\mathbf{e}_\emptyset$) by comparing it to the methodology of Desai et al.~\cite{desai2023hyperbolic}, who instead use the empirical mean of the embedding vector of train set items as the root position for CLIP. We use $\mathbf{\bar{e}}$ to indicate this empirical mean, calculating it as the (unit-normalized) mean of the embeddings of all texts in the \dataset{} train set (treating each caption from each tier, both positive and negative, as a separate text to embed). We do not use the corresponding images in this calculation since all our fine-tuning procedure uses only the textual data from \dataset{}. As is seen in the table, this leads to a degradation in hierarchical performance as reflected in poorer \dataset{} recall relative to our original fine-tuning method. We also consider two additional alternatives: Using the origin ($\mathbf{0}$) as a fixed root, and by using the original embedding $\mathbf{e}_\emptyset$ of the empty string as a fixed root vector (i.e. not allowing its position to change during fine-tuning). As is seen in the table, the initialization of the root has a critical effect, as the model fails by most hierarchical understanding metrics when the root is fixed at $\mathbf{0}$, while when the root is instead fixed at the original embedding $\mathbf{e}_\emptyset$, performance is much closer to our original fine-tuning procedure.

Conversely, we also perform an ablation in which we learn the root embedding position alone while keeping all other embeddings fixed (``only learning root'' in Table \ref{tab:supp_abl}). We see that this yields performance close to the base CLIP$^\text{B}$ model without fine-tuning, demonstrating that the root embedding position alone is insufficient for our fine-tuning framework.

We also ablate the effect of our regularization loss by fine-tuning with $\lambda_{reg} = 0$ (see Section \ref{sec:suppl_hyper}).  As is seen in the table, this loss is needed to avoid a drop in performance on a standard multimodal task (COCO text$\to$image retrieval). In other words, the regularization loss prevents catastrophic forgetting of pretraining knowledge.

To show the impact of the four-tiered structure of \dataset{}, we ablate by removing an intermediate tier (the second-most generic items) for fine-tuning. Results are listed as ``3 tiers'' in Table \ref{tab:supp_abl}. We see a particularly significant drop in performance on metrics such as $\tau_d$ for \dataset{} and performance on HyperLex, which are heavily weighted towards understanding lexical entailment and the relative semantics of generic items. This matches the observation that VLMs are known to struggle with more generic concepts, as discussed in the main paper, and hence the more generic tiers provide particularly important supervision during our fine-tuning process.

We ablate the effect of the positive-negative example pairs used during fine-tuning. We ablate the use of negative examples by setting all loss terms corresponding to negative pairs (i.e. $\xineg$, where $(\mathbf{e}, \mathbf{e}', \mathbf{e}'')$ is a caption triplet as described in the main paper) to zero during training.
As is seen in the table, this degrades the quality of representations across tasks, particularly hierarchical understanding tasks.

Finally, we test two alternatives to our fine-tuning method. We test using the original CLIP contrastive objective (instead of our \fwk{} loss) using text labels randomly sampled from from each hierarchy. This mostly performs similarly to pretrained CLIP, far under-performing our \fwk{} fine-tuning across tasks and thus suggesting that our results cannot be attributed only to learning the general semantics of the \dataset{} dataset. In addition, we test training from scratch with our loss rather than using pretrained CLIP weights. We initialize the CLIP-base architecture randomly (and remove the regularization loss minimizing deviation from the initial weights). As our training only adjusts the text encoder, this exhibits random performance on image matching tasks (\dataset{} P \& R, BREEDS, COCO). On text-only tasks we find that it has learned some hierarchical understanding, while mostly underperforming pretrained CLIP. This strengthens our core motivation of probing and optimizing the emergent geometry of pre-trained VLMs, which evidently have already learned embedding configurations which are highly compatible with visual-semantic hierarchies.

\subsection{Effect of Text Length}

We note that text length may be partially correlated with specificity or generality, as seen in Figure \ref{fig:supp_inacc} with more specific tiers of \dataset{} generally being longer. To test whether our \fwk{} fine-tuning results are mainly attributable to text length, we examine the performance of CLIP$_{\text{FT}}^\text{B}$ on \dataset{} when controlling for text length. We do this by comparing our overall results to those when restricting to controlled subsets of the \dataset{} test set -- when either the first two (most general) tiers have the same word count (72\% of the test data), or when the last two (most specific) tiers differ by at most four words (34\% of the test data). In both cases, we find all metrics (P, R, $\tau_d$) to differ by less than 0.015 in value with their values on the entire test set. We also note that the external HyperLex dataset consists of single words alone, and our results show a significant improvement on this benchmark after fine-tuning. Overall, we conclude that, while text length may serve as a cue to specificity, the observed hierarchical effects in the VLMs we investigate are not primarily due to text length.
\subsection{Additional Cross-Modal Task Results} \label{sec:supp_xmodal}

\begin{table*}[t]
  \centering
  \begin{tabularx}{0.85\textwidth}{lccccccccccc}
    \toprule
    &
    \multicolumn{3}{c}{COCO} & &
    \multicolumn{3}{c}{Flickr30K} & &
    CIFAR-10 & &
    CIFAR-100
    \\
    \cmidrule(lr){2-4}
    \cmidrule(lr){6-8}
    \cmidrule(lr){10-10}
    \cmidrule(lr){12-12}
    Model &
    R@1 & R@5 & R@10
    & & R@1 & R@5 & R@10
    & & acc
    & & acc
    \\
    \midrule
$\text{CLIP}^{\text{B}}$ & 0.30 & 0.55 & 0.66 &  & 0.59 & 0.83 & 0.90 &  & 0.90 &  & 0.66 \\
$\text{CLIP}^{\text{B}}_{\text{FT}}$ & 0.31 & 0.56 & 0.67 &  & 0.60 & 0.84 & 0.91 &  & 0.90 &  & 0.65 \\
\midrule
$\text{CLIP}^{\text{L}}$ & 0.36 & 0.60 & 0.70 &  & 0.65 & 0.87 & 0.92 &  & 0.96 &  & 0.78 \\
$\text{CLIP}^{\text{L}}_{\text{FT}}$ & 0.36 & 0.61 & 0.71 &  & 0.66 & 0.88 & 0.93 &  & 0.96 &  & 0.79 \\
\midrule
$\text{OpenCLIP}^{\text{B}}$ & 0.39 & 0.65 & 0.75 &  & 0.67 & 0.88 & 0.93 &  & 0.94 &  & 0.76 \\
$\text{OpenCLIP}^{\text{B}}_{\text{FT}}$ & 0.39 & 0.65 & 0.74 &  & 0.67 & 0.88 & 0.93 &  & 0.93 &  & 0.77 \\
\midrule
$\text{OpenCLIP}^{\text{H}}$ & 0.48 & 0.73 & 0.81 &  & 0.78 & 0.94 & 0.97 &  & 0.98 &  & 0.86 \\
$\text{OpenCLIP}^{\text{H}}_{\text{FT}}$ & 0.48 & 0.73 & 0.81 &  & 0.77 & 0.94 & 0.97 &  & 0.98 &  & 0.86 \\
\midrule
$\text{ALIGN}$ & 0.42 & 0.67 & 0.77 &  & 0.74 & 0.92 & 0.96 &  & 0.78 &  & 0.53 \\
$\text{ALIGN}_{\text{FT}}$ & 0.42 & 0.67 & 0.77 &  & 0.74 & 0.92 & 0.96 &  & 0.78 &  & 0.53 \\
    \bottomrule
  \end{tabularx}
  \caption{Results on standard cross-modal tasks before and after fine-tuning. Superscript letters indicate model size. \emph{acc} indicates categorical accuracy, and R@k refers to recall at k. As seen above, our fine-tuning has a negligible impact on performance on these tasks.}
\label{tab:supp_crossmodal}
\end{table*}

Table \ref{tab:supp_crossmodal} shows performance on additional standard cross-modal tasks and metrics, along with those reported in our main paper. These include text$\to$image retrieval on MS-COCO~\cite{lin2014microsoft,chen2015microsoft} (val set) and Flickr30K~\cite{young2014image} (test set), and zero-shot image classification on CIFAR-10 and CIFAR-100~\cite{Krizhevsky09learningmultiple}. For classification tasks, we use the prompts from Radford et al.~\cite{radford2021learning} along with their procedure for aggregating scores across prompts. Results are shown in Table \ref{tab:supp_crossmodal}. As is seen there, our fine-tuning procedure has a negligible impact on performance across these tasks (while boosting hierarchical understanding, as described in the main paper).
\subsection{Additional Qualitative Results}

Additional qualitative results can be viewed using our interactive visualization tool; see Section \ref{sec:supp_viz}.

\begin{figure}
\centering
    \jsubfig{
        \includegraphics[height=0.20\linewidth]{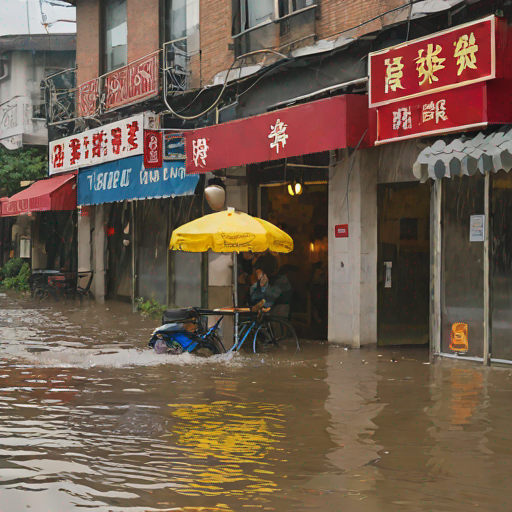}
    }{\tiny{disaster → flood → \hl{person taken away} during flood → a not so lucky chinese take away in the unprecedented floods}}
    \hfill
    \jsubfig{
        \includegraphics[height=0.20\linewidth]{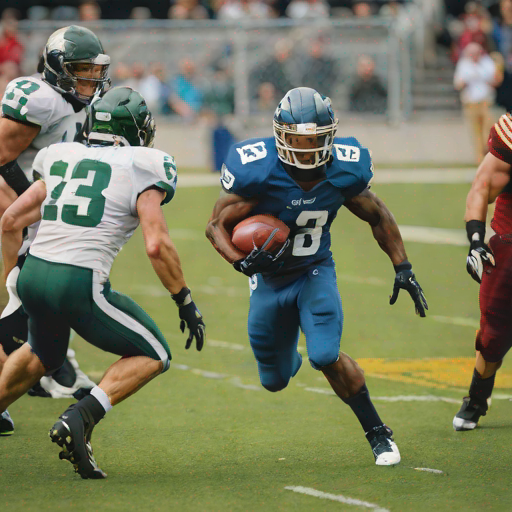}
    }{\tiny{\hl{yardage} → \hl{playing yards} → american football player runs for yardage against sports team. → american football player runs for yardage against sports team during the game .
    }}
    \hfill
    \jsubfig{
        \includegraphics[height=0.20\linewidth]{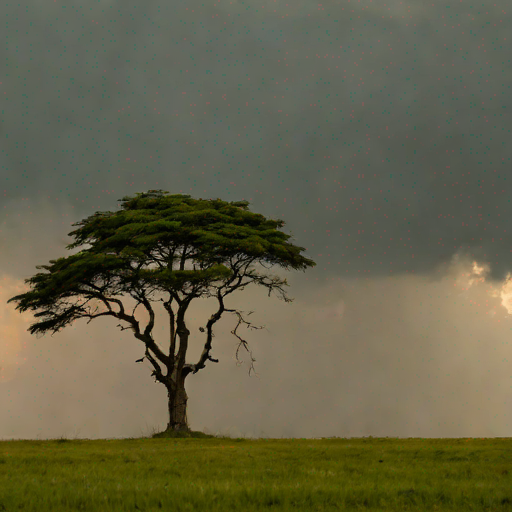}
    }{\tiny{\hl{solitary} → solitary tree → a solitary tree on the edge of the storm → a solitary tree on the edge appears to anticipate an oncoming storm}}
    \hfill
    \jsubfig{
        \includegraphics[height=0.20\linewidth]{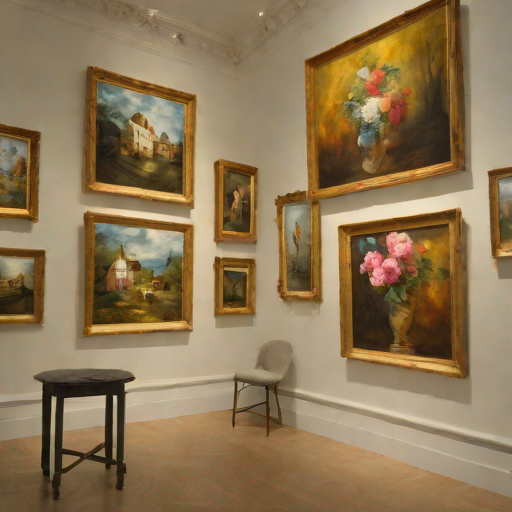}
    }{\tiny{\hl{limited} → \hl{limited 2} exhibition → \hl{limited 2} artwork show → \hl{limited 2} art gallery : a selection of stunning \hl{we showcase}}}
    \vspace{-10pt}
    \caption{
   Automatically-generated hierarchies from the HierarCaps train set, with inaccuracies \hl{highlighted}. We show similar synthetic images rather than the original images due to licensing.
   }
    \label{fig:supp_inacc}
\end{figure}

Browsing the uncurated \dataset{ train set illustrates the automatically-produced caption hierarchies in this benchmark, which we manually curated in the test set. This was necessary due to various types of noise which could occur as a result of the automatic generation process; we illustrate representative examples of this in Figure \ref{fig:supp_inacc}. As is seen there, such noise may include include minor misunderstanding of wording on the part of the language model (image 1), the use of abstract terms which are not incorrect per se but do not concretely describe the image (images 2--3), and noise inherited from the long image captions in the last tier which come from an existing captioning dataset (image 4). These are often attributable to the limited capacity of the language model used to produce these captions and the production of caption hierarchies with a text-only pipeline, both drawbacks of the efficient use of the small distilled text-only language model $\mathcal{H}$.}
\subsection{Comparison to Entailment Cones} \label{sec:supp_ec_comparison}

\begin{table}[t]
  \centering
  \begin{tabularx}{0.28\linewidth}{lccc}
    \toprule
    &
    \multicolumn{3}{c}{\dataset{}}
    \\
    \cmidrule(lr){2-4}
    Model &
    P & R & $\tau_d$
    \\
    \midrule
    $\text{CLIP}^{\text{B}}$ & 0.14 & 0.36 & 0.89
    \\
    $\text{CLIP}^{\text{B}}_{\text{FT}}$ & 0.15 & 0.47 & 0.99
    \\
    \midrule
    $\alpha=0$ & 0.13 & 0.30 & 0.97 \\
    $\alpha=\pi/8$ & 0.14 & 0.33 & 0.97 \\
    $\alpha=\pi/4$ & 0.14 & 0.35 & 0.98 \\
    $\alpha=\pi/2$ & 0.17 & 0.46 & 0.99 \\
    \bottomrule
  \end{tabularx}
  \caption{Results when fine-tuning with EC-based losses. We first reproduce our results for a CLIP model before and after fine-tuning using our full framework including a \fwk{} loss. We then report results when substituting an EC loss for our contrastive \fwk{} loss, as described in Section \ref{sec:supp_ec_comparison}. In particular, we see increased performance as the margin $\alpha$ increases, which we show (in Section \ref{sec:supp_ec_deriv}) has the \fwk{} loss as its limiting case.}
\label{tab:supp_ec}
\end{table}

We compare our \fwk{} framework to the Entailment Cone (EC) framework of Ganea et al.~\cite{ganea2018hyperbolic} by replacing the loss $\mathcal{L}_{\fwk{}}$ in fine-tuning with an EC loss. In particular, we consider margin EC losses of the form
$$\mathcal{L}_{EC}^{(\alpha, \epsilon)} := \left[\pm (\xipos - \theta_\epsilon + \alpha)\right]^+$$

\noindent where $\left[x\right]^+ = \max(0, x)$ is the positive part (ReLU) function, $\alpha$ is the chosen margin, and $\pm$ is positive for positive pairs and negative for negative pairs. The entailment cone half-angle $\theta_\epsilon$ is defined by $\sin \theta_\epsilon = \min(1, \epsilon / d_{\mathbf{r}}(\mathbf{e}))$. We use this loss by replacing $\xipos$ and $\xineg$ in the calculation of $\mathcal{L}_{\fwk}$ with the expression above, using fixed $\epsilon = 0.05$, and proceeding to aggregate losses across tiers and samples as described in Section \ref{sec:suppl_hyper}.
See Table \ref{tab:supp_ec} for results using various values of $\alpha$, beyond those reported in the main paper. From these results we see improved performance as the margin $\alpha$ increases, motivating the derivation of the \fwk{} loss from this loss as $\alpha \to \infty$, as described in Section \ref{sec:supp_ec_deriv}.
\subsection{Comparison to Existing HyperLex Methods}

\begin{table}[t]
  \centering
  \begin{tabularx}{0.28\columnwidth}{lcc}
    \toprule
    \textbf{Model} & $\rho_{all}$ & $\rho_N$ \\
    \midrule
    \textbf{Ours} \\
    CLIP$^\text{B}$ & 0.44 & 0.48 \\
    CLIP$^\text{B}_\text{FT}$ & 0.51 & 0.55 \\
    \midrule
    \textbf{Dedicated} \\
    LEAR~\cite{vulic2017specialising} & 0.69 & 0.71  \\
    DOA-E~\cite{athiwaratkun2018hierarchical} & - & 0.59 \\
    HyperVec~\cite{nguyen2017hierarchical} & 0.54 & - \\
    Poincaré~\cite{nickel2017poincare} & - & 0.51 \\
    PARA ~\cite{mrkvsic2016counter,nguyen2017hierarchical} & 0.32 & - \\
    FR~\cite{vulic2017hyperlex} & 0.28 & 0.28 \\
    \bottomrule
  \end{tabularx}
  \caption{Results on the HyperLex lexical entailment benchmark, comparing our results when probing CLIP (before or after model alignment) to existing dedicated methods. We use the following abbreviations: Frequency Ratio (FR), PARAGRAM (PARA), Order Embeddings (OE). We reproduce values from the cited works, noting that some only report one of $\rho_{all}$ or $\rho_N$. We see that our method is competitive with existing dedicated methods for this task.}
\label{tab:supp_hyperlex}
\end{table}

See Table \ref{tab:supp_hyperlex} for a comparison of our results on HyperLex (using a CLIP model as-is as well as after our fine-tuning), along with various dedicated methods for learning lexical entailment. We reproduce the $\rho_{all}$ and $\rho_N$ metrics reported by each work -- the Spearman correlation between predicted scores and ground-truth entailment scores from HyperLex, for all items and for nouns respectively. These results show that our method is competitive in this field; while we do not surpass the state-of-the-art for dedicated methods, we do show that CLIP can achieve performance of a similar order of magnitude to existing methods, either when used zero-shot or with model alignment (fine-tuning). We also emphasize the significant effect of model alignment, which improves both $\rho_{all}$ and $\rho_N$ by $0.07$ and reduces the gap between CLIP and the highest-performing dedicated models.
\subsection{BREEDS Single Prompt Baseline Results}

\begin{table*}[t]
  \centering
  \begin{tabularx}{0.68\textwidth}{lcccc}
    \toprule
    &
    \multicolumn{2}{c}{CLIP$^\text{B}$} &
    \multicolumn{2}{c}{CLIP$^\text{B}_{\text{FT}}$} \\
    \cmidrule(lr){2-3}
    \cmidrule(lr){4-5}
    Prompt & R@1 & R@5 & R@1 & R@5 \\
    \midrule
    \texttt{a picture of a \{\}, a type of \{\}} &
    0.00 & 0.02 & 0.00 & 0.02 \\
    \texttt{a \{\} of type \{\}} &
    0.01 & 0.04 & 0.01 & 0.03 \\
    \texttt{this is a \{\}, a type of \{\}} &
    0.01 & 0.03 & 0.00 & 0.03
    \\
    \texttt{a picture of a \{\} of type \{\}} &
    0.01 & 0.04 & 0.01 & 0.03
    \\
    \texttt{a picture of a \{\} which is a \{\}} &
    0.00 & 0.03 & 0.00 & 0.02
    \\
    \midrule
    (ours) & 0.22 & 0.50 & 0.24 & 0.55 \\
    \bottomrule
  \end{tabularx}
  \caption{Baseline results for classification of BREEDS label pairs using a single prompt containing two slots for labels. In the above prompts, \texttt{\{\}} indicates the slot for inserting a label text. We also reproduce results for our proposed inference method from the main paper at the bottom. We see that the baseline yields near-random performance and is significantly outperformed by our proposed method.}
\label{tab:supp_breeds_baseline}
\end{table*}

To provide a baseline comparison and provide justification for our inference method, we evaluate performance on BREEDS when predicting label pairs inserted into a single prompt. We use the same BREEDS subsets and labels as discussed in the main paper. To predict a pair of labels (x, y), we insert both labels into a prompt with two slots (e.g. \texttt{a \{\} of type \{\}}, where \texttt{\{\}} indicates a slot for inserting text), and yield the pair of distinct labels (x, y) which maximizes the embedding similarity of the given prompt with the image. In Table \ref{tab:supp_breeds_baseline}, we report performance for various such prompts on a CLIP model evaluated both before and after model alignment; we also reproduce the performance of our probing method from the main paper. As seen in the table, zero-shot classification of paired hierarchical labels using this single prompt-based inference method yields near-random performance, unlike our proposed inference method.

\end{document}